\documentclass[10pt,twocolumn,letterpaper]{article}

\usepackage[pagenumbers]{cvpr} % To force page numbers, e.g. for an arXiv version

\usepackage{mathtools}
\usepackage{xcolor}
\usepackage{float} % To manage positioning
\usepackage{caption} % To control caption widths
\def\paperID{10829} % *** Enter the Paper ID here
\def\confName{CVPR}
\def\confYear{2024}

\usepackage{graphicx}
\usepackage{amsmath}
\usepackage{amssymb}
\usepackage{booktabs}
\usepackage{microtype}
\usepackage{adjustbox}
\usepackage{listings}
\usepackage{pythonhighlight}
\usepackage{float}
\usepackage{tablefootnote}
\usepackage{tabularx}
\usepackage{wrapfig}
\usepackage{enumitem}
\usepackage[ruled, vlined]{algorithm2e}

\definecolor{cvprblue}{rgb}{0.21,0.49,0.74}
\usepackage[pagebackref,breaklinks,colorlinks,citecolor=cvprblue]{hyperref}

\usepackage[capitalize]{cleveref}
\crefname{section}{Sec.}{Secs.}
\Crefname{section}{Section}{Sections}
\Crefname{table}{Table}{Tables}
\crefname{table}{Tab.}{Tabs.}

\title{Common Sense Bias Modeling for Classification Tasks}

\author{Miao Zhang$^{1}$ \qquad Zee Fryer$^{2}$ \qquad Ben Colman$^{2}$ \qquad Ali Shahriyari$^{2}$ \qquad Gaurav Bharaj$^{2}$ \\
{$^{1}$New York University \qquad  $^{2}$Reality Defender Inc} \\
{\tt\small miaozhng@nyu.edu}, 
{\tt\small \{zee, ben, ali, gaurav\}@realitydefender.ai}
}
\date{}

\begin{document}
\maketitle

\begin{abstract}
Machine learning model bias can arise from dataset composition: correlated sensitive features can disturb the downstream classification model's decision boundary and lead to performance differences along these features. Existing de-biasing works tackle most prominent bias features, like colors of digits or background of animals. However, a real-world dataset often includes a large number of feature correlations, that manifest intrinsically in the data as common sense information. Such spurious visual cues can further reduce model robustness. Thus, practitioners desire the whole picture of correlations and flexibility to treat concerned bias for specific domain tasks. With this goal, we propose a novel framework to extract comprehensive bias information in image datasets based on textual descriptions, a common sense-rich modality. Specifically, features are constructed by clustering noun phrase embeddings of similar semantics. Each feature's appearance across a dataset is inferred and their co-occurrence statistics are measured, with spurious correlations optionally examined by a human-in-the-loop interface. Downstream experiments show that our method discovers novel model biases on multiple image benchmark datasets. Furthermore, the discovered bias can be mitigated by a simple data re-weighting strategy that de-correlates the features, and outperforms state-of-the-art unsupervised bias mitigation methods.

% and the intervention using discovered feature distribution outperforms state-of-the-art unsupervised bias mitigation methods.

% A common sense reasoning approach to bias detection that discovers ``concept clusters'' and their correlations. 

\end{abstract}

\section{Introduction}
\label{sec:intro}

% https://docs.google.com/drawings/d/1KeEOd0l9Uug-VBYrH7OuMqivaZ8Sgg1hxh5D0soBDEA/edit?usp=sharing
% \begin{figure}[!htbp]
%   \centering
%     \includegraphics[width=1\linewidth]{figures/Teaser.pdf}
%     \caption{Sensitive features are everywhere: Small objects which co-occur frequently with the target can affect model prediction, for example, kites and people in MS-COCO images. Sensitive features like this are of multiple types and may cause different downstream biases.
%     %Existing  work often struggles to target and treat bias when the feature groups are not easily distinguishable in latent space.
%     Our method aims to discover comprehensive sensitive correlations in a dataset based on common-sense descriptions, and treat biases which have not been explored in literature.
%     }
%     \vspace{-3mm}
%   \label{fig:teaser}
% \end{figure}

% to add
% Caption may come from human or LLM

\begin{figure}
    \centering
    \includegraphics[width=0.48\textwidth]{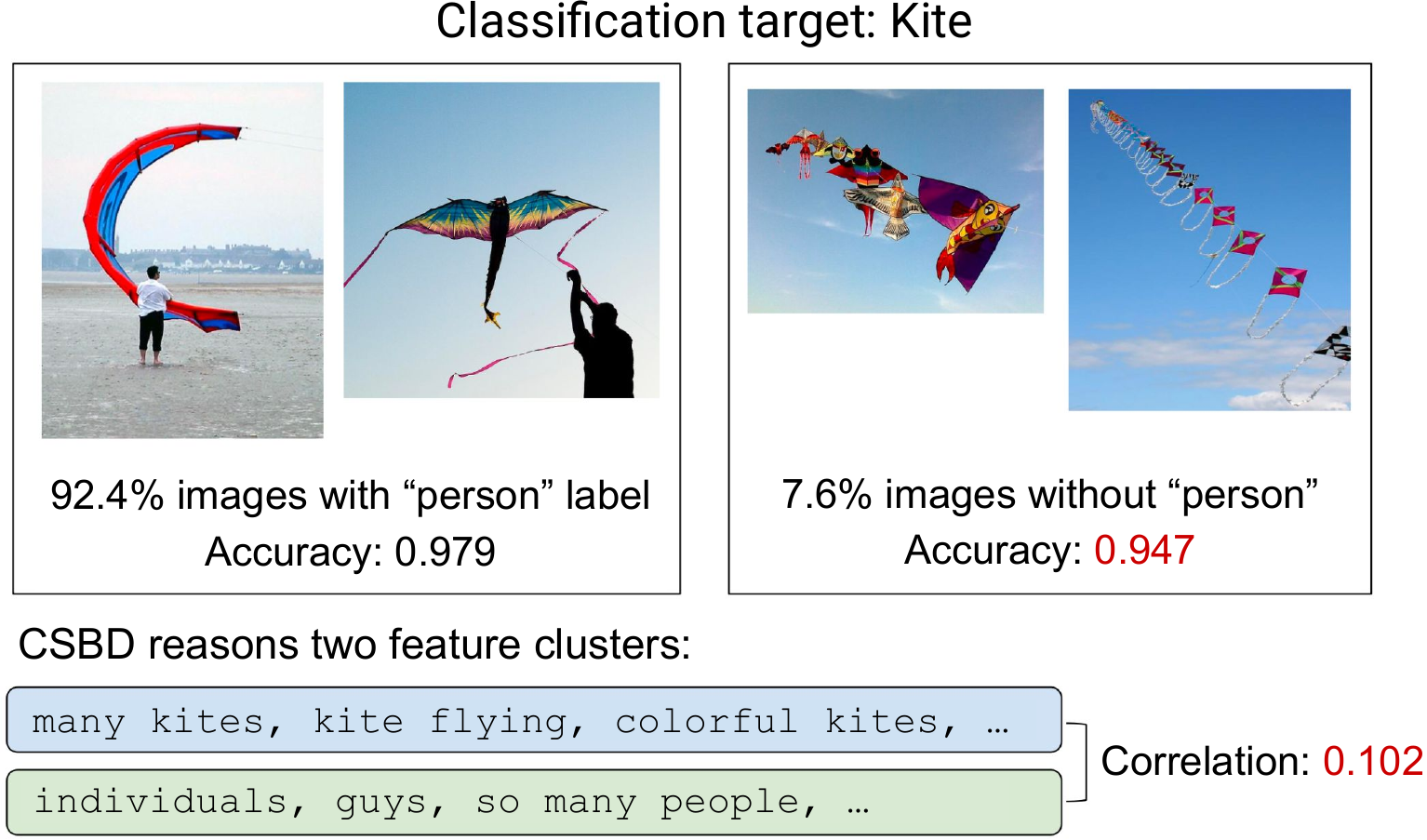}
    \caption{Spurious features are everywhere: Objects which co-occur frequently with the target can affect model prediction, for example, kites and people in MS-COCO. Spurious features like this are of multiple types and may cause different downstream biases.
    %Existing  work often struggles to target and treat bias when the feature groups are not easily distinguishable in latent space.
    Our method aims to discover a comprehensive list of them based on common sense descriptions, and treat biases which have not been explored in literature.
    }
  \label{fig:teaser}
\end{figure}

% Sentence 1: CONTEXT - why now?
Computer vision has been deployed with dramatically more diverse data (both realistic and generative) in recent years, which has drawn attention to data cleaning and bias reduction~\cite{torralba2011unbiased}. %built-in bias 
One common bias is due to the frequent co-occurrence of target features with context features (illustrated in Figure \ref{fig:teaser}), that downstream task models may rely on to make predictions even though the relationship is not robust and generalizable~\cite{singh2020don, basu2023inspecting, wang2024navigate}. 

Current unsupervised algorithms detect bias informed by model performance. For example, to separate samples aligning with and conflict to bias in model latent space ~\cite{sohoni2020no, krishnakumar2021udis, ahn2022mitigating, li2022discover, seo2022unsupervised}, or by model gradient~\cite{ahn2022mitigating}. We observe that most approaches have only tackled limited  bias types with prominent and delicate features, like the color of digits, the background of animals~\cite{nam2020learning, sagawa2019distributionally}, or the gender of the person doing certain activities~\cite{zhao2017men}. It remains to be an open question whether more general image features, including those coarse and subtle ones in a dataset will correlate with the target feature in an \emph{unwanted} way and affect model predictions. For example, small objects like a keyboard nearby a cat, a person nearby a kite which we discover to compose short-cut learning (Figure~\ref{fig:teaser}), have been overlooked for bias modeling.

% However, the approaches are limited to the most amplified features and may not generalize to discovering multiple existing bias. For example, the kite classifier in \cref{fig:teaser} is affected by the spurious feature of person presence, but the existing discussion in literature is limited background.  

% previous studies have shown that visual features that occupy smaller spatial area can obtain less mo~\cite{truong2023fredom}. Subtle facial features such as eyeglasses frames or small object features like a keyboard nearby a cat may be overlooked for bias modeling.

% We aim to circumvent the need for latent space operation and instead to use natural language to discover sensitive attribute distribution from data, therefore be able to fix performance disparity by light-weighted intervention. The approach also provides transparency and interpretability to the cause of unfairness.

% These methods rely on image-based embeddings and thus are limited to only identifying certain kinds of bias that could be captured in that embedding space. 
These observations suggest to enrich bias signals, and we propose to model with additional data modalities, specifically text descriptions of images. Differentiating to labels, descriptions can cover a wider range of features that humans perceive (with respect to whole dataset instead of individual data samples as descriptions can be noisy). We show by experiments that debiasing based on supervision from description-derived bias obtains state-of-the-art results.

% for example generated due to shadows
% These methods are limited in the ``kind'' of bias they are able to detection -- contextual example of "shadow vs hair" or "white coat and male doctor".

% While in text sense, such features are equally important as the overall facial structure w.r.t to feature importance in bias analysis. 

% New explorations are done in this direction for leveraging  for model bias and error detection: 

``Common sense reasoning'' by natural language~\cite{lake2017building, antol2015vqa, radford2021learning, diomataris2021grounding} is a popular approach for high-level abstract understanding of images and reasoning for subtle image features. There have been some early explorations in using this approach for bias discovery. For example, studies encode verbalized spurious features into text embeddings, then aligning to model latent space to measure feature influence~\cite{wu2023discover, zhang2023diagnosing}. 
% and the model's sensitivity to each feature can be quantified by aligning the embedding to the model's inner state~\cite{wu2023discover, zhang2023diagnosing}. 
However, the use cases of the existing common sense bias modelings are restricted 
by the reliance on multi-modal models like CLIP~\cite{radford2021learning} or generative models~\cite{rombach2022high} to align embedding space of image and text. This additional step of alignment is challenging: it has a high requirements for representation quality and may not capture features from fine-grained image regions.
% in two major aspects: First, though prompt-encoded features can be more diverse than traditional image-based patterns, they are limited by the prior knowledge of prompt writers and cannot expose unlabeled features\footnote{E.g. the MS-COCO dataset is only labeled for 80 objects, but many other unlabeled objects also appear in the images.} in a dataset. 
% Second, the additional step of multi-modal embedding space alignment using models like CLIP~\cite{radford2021learning} or generative models~\cite{rombach2022high} lacks the control to align fine-grained image regions with text, and may inherit extra bias from the models~\cite{bommasani2021opportunities, schramowski2022large}. 
There are also studies that analyze bias in dataset captions~\cite{van2016stereotyping, wang2022measuring}, but their focus is on linguistic stereotypes within captions themselves, rather than using the captions to debias visual recognition. 

% Sentence 2: NEED - why does the reader care?
% Importance of bias discovery\footnote{\url{https://www.theverge.com/2018/1/12/16882408/google-racist-gorillas-photo-recognition-algorithm-ai}} rely on the image-based embedding or human defined labels for bias, while bias due to contextual reasoning and "second-degree" relations does not exists. 

% Sentence 3: TASK - what do we do?
To explore this direction, we design a novel description-based reasoning method for feature correlation and downstream model bias, called Common Sense Bias Discovery (CSBD). Given a description corpus of images (for example, captions), we unpack it into ``features'' through noun chunk clustering. These features are extracted from the semantic components of descriptions, which represent various image content, not limited by prior knowledge or cross-modal alignments. Then, correlations between features are quantified based on their co-occurrence across all data samples. It should note that discovered correlated features are not necessarily bias: for example, a correlation between ``\texttt{teeth}'' and ``\texttt{smile}'' in a dataset of face images is both expected and entirely benign. Therefore, we maintain a human-in-the-loop component to identify spurious correlations that should be addressed for specific tasks.  However, following the identification step, our method does not require human intervention to mitigate the downstream model bias caused by these correlations. Our contributions are as follows: 

% capture high-level abstract understanding of the image content, and are less affected by low-level feature shifts like brightness, resolutions, etc ~\cite{majumdar2021unravelling}. Then, 

% The process is performed in data context, motivated by the advantage of improving data quality than constraining model bias to avoid predictive accuracy sacrifice ~\cite{chen2018my}.

% To the best of our knowledge, no system for model-agnostic text-based reasoning for bias detection exits.

% Why human expertise if incorporated into the pipeline. will have the flexibility to how broad the category is. 

% Sentence 4: OBJECT - what does this document do?

% Sentence 5: FINDINGS - what did we discover?

% Sentence 6: CONCLUSIONS - so what?

% Sentence 7: PERSPECTIVES - what now?

% We present a list of unwarranted correlations inferred from image descriptions.

\begin{enumerate}
    \item A novel framework that uses textual descriptions obtained via humans or vision-language models that: (i) comprehensively analyzes spurious correlations contained in image datasets, and (ii) provides guidance on debiased downstream model training. Specifically: 
    \item A common sense reasoning approach that generates human interpretable ``feature clusters''. Based on the discovered clusters, the formulations to derive pairwise correlations and re-sampling weights for model debiasing.
    
    % \item Experimental results show the biases discovered by our approach (w.r.t. static sensitive features on multiple benchmarks), to the best our knowledge has have not been previously addressed. While the resulting mitigation achieves state-of-the-art performance.

    \item Experimental results show that our approach discovers biases with respect to static spurious features on multiple benchmarks, which to the best our knowledge have not been previously addressed. The resulting mitigation achieves state-of-the-art performances.
\end{enumerate}

% proposed by our reasoning approach

% We perform experiments on two datasets, CelebA and MS-COCO, and show that the correlations we identify do result in training biased classifiers for relevant downstream tasks; for example, smiling detection performance on CelebA differs among gender groups, and cat detection on MS-COCO is affected by the presence of a couch or keyboard in the image. Having exposed such correlations in the data, we mitigate the bias using a simple re-weighting strategy.

% We continue the effort in identifying human-interpretable biased attributes with an automated pipeline, besides human-in-the-loop for judging harmfulness of the bias. They require common sense human knowledge which cannot be accomplished by computer.

% \gb{Image feature extraction can overlook subtle feature such as thin eyeglasses or faint facial attributes such as a smile, while in text sense, such features are equally important as the overall facial structure w.r.t to feature importance.}

\section{Related Work}
\label{sec:rw}

% \subsection{Bias discovery}
\paragraph{Unsupervised bias discovery.} Many recent studies have improved model robustness without requiring sensitive feature annotation. One approach is to assume that easy-to-learn data samples lead to shortcut feature learning, thus resulting in biased classifiers. These bias-aligning samples have higher prediction correctness and confidence~\cite{kim2021biaswap,liu2021just, zhang2022correct, li2023partition}, larger gradient~\cite{ahn2022mitigating}, or being fitted early in training~\cite{nam2020learning, lee2023revisiting}. Feature clustering has also been used, which leverages the observation that samples with same the feature are located closely in model latent space. Bias is then identified based on the trained model's unequal performance across clusters~\cite{seo2022unsupervised, krishnakumar2021udis}. Another approach finds bias related subgroups based on a certain amount of latent space directions that are highly correlated to model performance~\cite{zhang2024discover}.
% model attributes as latent variables and identify influential attributes by latent space distribution of training samples~\cite{amini2019uncovering, lang2021explaining}, 
% or find subsets of samples which have maximum predicted probabilities on a target to indicate a biased group ~\cite{li2022discover}. 
Most methods rely on additional biased learning to infer discounted samples, which might not be robust to shifts in training algorithm or schedules. Also, the implicit bias discovery used in these methods lacks transparency and interpretability. We propose a novel direction to discover dataset bias by description reasoning, which is agnostic and generalizable to different downstream learning schemes.

\paragraph{Bias mitigation.}
Methods that identify bias-conflicting samples or subgroups use them as supervision to upweight low performers, thus training a debiased model~\cite{nam2020learning, krishnakumar2021udis, liu2021just, ahn2022mitigating, seo2022unsupervised, lee2023revisiting, zhang2024discover}. Alternatively, methods encourage models to learn similar representations for samples with the same target but different sensitive features via contrastive learning~\cite{zhang2022correct, park2022fair, zhang2022fairness}, e.g., augmenting bias-aligned images to alter sensitive features while maintaining the target content~\cite{kim2021biaswap, ramaswamy2021fair, lim2023biasadv}. Bias mitigation method using regularization loss penalizes models for violations of the Equal Opportunity fairness criterion~\cite{li2022discover}. 
% or to update latent variables of a variational autoencoder that debiasing is based on~\cite{amini2019uncovering}. 
These frameworks address bias identified during model learning process and perform interventions accordingly. We approach the problem from a new prospective, recognizing that bias can take various forms captured by the common sense information of a dataset. Analyzing feature connections from this information can lead to simple and effective bias mitigation.

% which is not explicitly annotated/implicit but

% cannot be generalized easily to different learning protocols

% \subsection{Text-Image methods}
% \paragraph{Visual-semantic Grounding.} To ground representations of images with natural linguistic expression, models learn an embedding space where semantically similar textual and visual features are well aligned~\cite{karpathy2015deep, wu2019unified, jia2021scaling, radford2021learning}. Such models can be extended to align representations from local image regions, and can adapt to various multi-modal tasks ~\cite{yuan2021florence, alayrac2022flamingo, zhang2022glipv2, zhong2022regionclip}. They motivate new downstream applications via ``prompting''; examples include domain adaptation by augmenting image embeddings prompted by target domain text descriptions~\cite{dunlap2022using},  aligning positional features of 
% % zee: todo
% animals' pose estimation using pose-specific text prompts~\cite{zhang2023clamp}, and conditional image synthesis by training token generators for knowledge transfer~\cite{sohn2023visual}. Unlike methods that leverage multi-modal embedding spaces to intervene model training, our analysis and intervention are 
% data-focused for balancing feature distribution, so circumvents the need to handcraft or fine-tune prompts with alternate learning structures~\cite{khattak2023maple, sohn2023visual}.

\paragraph{Bias discovery with human knowledge.} Building machine learning systems with human judgement is crucial for user-centric and responsible goals~\cite{lake2017building, wang2019designing}. It enables domain expertise to be involved for improving real-world performances with minimal cost~\cite{wu2022survey}. For bias discovery purposes, researchers have analyzed text-based item sets in visual question answering dataset and use association rules to interpret model behaviors~\cite{manjunatha2019explicit}. Human is included to examine biases, including interpreting the semantic meaning of specific sensitive features~\cite{li2021discover} and analyzing bias information represented by the decomposition of sample subgroups in model latent space~\cite{zhang2024discover}. The studies motivate us for interpretable bias distillation. Image descriptions or captions are data formats rich of cognitive and common sense knowledge, thus are of great interests to be utilized for debiasing supervision. 

% However, such methods are limited by an empirical feature set which the model prediction bias is analyzed with. The human defined features may not generalize to different image domains, thus, we use clustering and correlation analysis for any datasets with an uncurated description corpus.

% Different to previous work which find spurious correlations in embedding space, we only use image embeddings to obtain target task relevance but analyze and resent common-sense correlations in the form of natural languages.
% generable and flexible. model-agnostic

\section{Method}
\label{sec:method}

% https://docs.google.com/drawings/d/1V7G8vdW_0XRoYkWqfd-2MunTXBc6qI0VNR73QheGTBw/edit
\begin{figure*}
  \centering
    \includegraphics[width=0.96\linewidth]{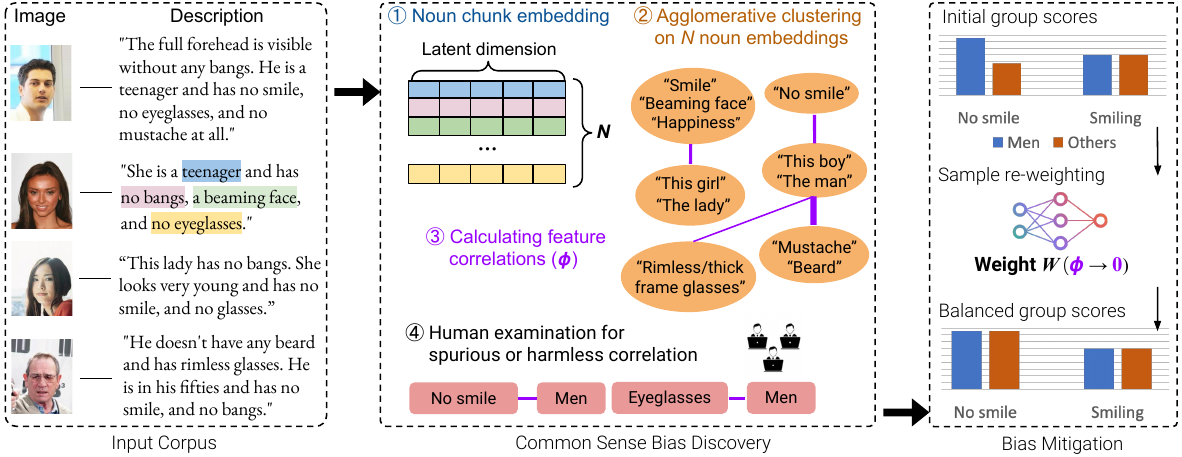}
    % \caption{\textit{Common Sense Bias Discovery (CSBD) system and the mitigation strategy overview}: Given the input corpus of image and corresponding description pairs (Left), descriptions are split into semantically meaningful noun chunks and encoded into representation vectors. A hierarchical clustering on the vector set generates a list of common sense feature clusters.  Then, correlations between every two feature distributions are computed and highly-correlated features are examined by a human for mitigation (Middle). Finally, 
    % %the target feature - sensitive feature correlations are judged as potential bias. They 
    % correlations are mitigated by adjusting image sampling weights during training, automatically calculated from the derived feature distributions (Right).}

    \caption{\textit{Common Sense Bias Discovery (CSBD) system and the mitigation strategy overview}: (Left) Input corpus of image and corresponding description pairs are given. (Middle) (1) Descriptions are split into semantically meaningful noun chunks and encoded into representation vectors. (2) A hierarchical clustering on the vector set generates a list of common sense feature clusters. (3) Correlations between the presence of every two features are computed. (4) Highly-correlated features are examined by a human for mitigation. (Right) Finally, correlations are mitigated by adjusting image sampling weights during model training, automatically calculated from the derived feature presence statistics.}
    
  \label{fig:overview}
\end{figure*}

% to add: description-agnostic method

As in Figure~\ref{fig:overview}, we present our framework of discovering and treating bias supervised by image descriptions, \textbf{C}ommon-\textbf{S}ense \textbf{B}ias \textbf{D}iscovery method, referred to as CSBD. 
In brief, given a corpus of (image, text description) pairs, our method generates clusters of noun chunks to indicate presence of features across samples. Then, pairwise feature\footnote{ ``Feature'' in this work refers to a cluster of semantically similar phrases, e.g.  ``\texttt{beaming smile}'' and ``\texttt{wide smile}'' are considered interchangeable and belong to one cluster/feature.} correlations are analyzed. This is followed by an optional human-in-the-loop module to allow a domain expert to review highly correlated features and select those that may negatively impact the downstream task. Finally, for any selected correlation to treat, we mitigate model bias via data re-weighting, without a requirement for sensitive group labels. Detailed algorithm steps are in the supplementary material. The three major components of the method are described in the following.

% Language-inferred feature distribution
% Discovering feature correlations
% Bias mitigation via re-weighting

\textbf{Language-inferred feature distribution.} Natural language image descriptions must first be split into information units to align with image objects. We use the spaCy ``en\_core\_web\_sm" model~\cite{honnibal2017spacy} to extract noun chunks (noun and adjective grouped together), e.g. obtaining ``\texttt{The girl}'' and ``\texttt{a big smile}'' from ``\texttt{The girl has a big smile}'', from all descriptions. 
% $M$ denotes the the total number of extracted noun chunks in a dataset. 
We then use a general purpose sentence embedding model, the Universal Sentence Encoder~\cite{cer2018universal}, to map the noun chunks into high-dimensional vectors that represent their semantics, denoted as $\boldsymbol{A} \in \mathbb{R}^{d}$. The selection of this text embedding model is based on accuracy and computing speed.
% Phrase vector dimensions are reduced to dimension $d$ (tuneable hyper-parameter) using PCA~\cite{jolliffe2002principal} before being clustered. The ablation results using different text-based embedding model and dimension reduction algorithm can be seen in the Supplementary.
Next, we implement agglomerative clustering on $\boldsymbol{A}$. It starts from single vector as a cluster and hierarchically merges pairs of clusters until a certain criteria is reached, given no prior knowledge of the cluster number. We set the criteria to be the maximum Euclidean distance between any vectors of two clusters, denoted as $z$. It is a tunable hyper-parameter that controls the granularity of the clusters, e.g. ``\texttt{a boy}'' and ``\texttt{a young man}'' may be clustered together with a larger $z$ and be in separate clusters with a smaller $z$. Each obtained cluster is viewed as a feature that is consistent with common sense descriptions. %is perceived with human knowledge.

\textbf{Discovering feature correlations.} Having extracted a set of feature clusters in the previous step, the next step is understanding feature co-occurence within the dataset. This allows us to identify spurious correlations relevant to the target task, which may cause model bias. 
% \gb{We assume a target task is pre-defined, on which image features that spuriously correlated to target attributes can result in biased model performance}. 
To quantify such correlations: first, we generate a binary indicator for whether each feature is present: $\boldsymbol{t}_{f} = [t_1, t_2, ..., t_N]$,  where $N$ is the size of the dataset, and $t_i=1$ if feature $f$ occurs in the $i^{th}$ image's description, otherwise $t_i=0$. Second, the phi coefficient $\phi$ \cite[p. 282]{Cramr1946MathematicalMO}\footnote{Also known as the Matthews correlation coefficient \cite{Matthews1975ComparisonOT}.}  is used to measure the association between every two feature indicators. The $\phi$ between two indicators $\boldsymbol{t}_f$ and $\boldsymbol{t}_{f'}$ is defined as follows:
%\footnote{$\neg$ means element-wise ``not''}
\begin{equation}\label{eq:phi def}
\small
\phi_{\boldsymbol{t}_{f}\boldsymbol{t}_{f'}} = \frac{x_{11}x_{00} - x_{10}x_{01}}{\sqrt{x_{1*}x_{0*}x_{*0}x_{*1}}},
\end{equation}
where $c \in [0, C)$, $f, f'\in [0, F_c]$, and
\begin{gather*}
\small
x_{11} = \|\boldsymbol{t}_{f} \cdot \boldsymbol{t}_{f'}\|_1, \quad 
x_{01} = \|(\neg \boldsymbol{t}_{f}) \cdot \boldsymbol{t}_{f'}\|_1, \\
\small
x_{10} = \|\boldsymbol{t}_{f} \cdot (\neg \boldsymbol{t}_{f'})\|_1, \quad
x_{00} = \|(\neg \boldsymbol{t}_{f}) \cdot (\neg \boldsymbol{t}_{f'})\|_1,\\
\small
x_{1*} = \|\boldsymbol{t}_{f}\|_1, \quad x_{0*} = \|\neg \boldsymbol{t}_{f}\|_1, \\ 
\small
x_{*0} = \|\neg \boldsymbol{t}_{f'}\|_1, \quad x_{*1} = \|\boldsymbol{t}_{f'}\|_1.
\end{gather*} 
$\neg$ denotes element-wise negation, $\| \cdot \|_1$ denotes the $L^1$ norm.

Two features are positively correlated (likely to co-occur in an image) if $\phi$ is a positive value, and are negatively correlated (rarely co-occur) if $\phi$ is a negative value. A $\phi$ near zero indicates two features which co-occur randomly. Intuitively, features that have high correlation with any target features can become shortcuts for model learning and cause biased decision-making toward certain subgroups~\cite{tian2022image, brown2023detecting}. Thus, feature pairs sorted by $\phi$ are returned for examination by humans, who have access to the common sense text description of each feature. 
% With the generated concept-feature hierarchy, we aim to discover \textit{sensitive feature correlations which affect task prediction}. First, we quantify model prediction's sensitivity to each feature's presence thus the task relevance, denoted as $S_{cf}$. The approach is to learn an activation vector for visual patterns of the feature, and use prediction's derivative with respect to directional change along the vector as sensitivity~\cite{kim2018interpretability}. The concept which has features with maximum sensitivity $\underset{c\in [0,C), f\in [0,F_c)}{max}S_{cf}$ is identified as the target concept $c'$ and it includes $F_{c'}$ target features. 
% % and the concepts model predictions are not sensitive to ($S_{cf} < \mu$) are excluded from the following analysis. 
% To analyze co-occurrence distribution between each target feature and other features, we compute phi coefficient ($\phi$) to measure association for indicators as:

% As an example, the feature hierarchy and correlations discovered with threshold $z=0.05$ on CelebA dataset~\cite{jiang2021talk} is shown in Figure.
\textbf{Strength of human evaluated bias.} In many cases, the extracted feature pairs are naturally connected (e.g. ``\texttt{teeth visible}'' with ``\texttt{a smile}'', and most facial features with ``\texttt{the face}''). These correlations are usually robust and generalizable, and thus viewed as benign. 
% zee: todo
The benign correlations usually exist, besides, the function of correlations vary across different downstream learning scenarios and not all correlations need to be treated. Therefore, a human-in-the-loop step is necessary to ensure that only spurious correlations that may impact the target task are selected. 
% Due to the presence of these benign correlations, and considering that the function of feature correlations vary across different downstream learning scenarios and not all correlations need to be treated, we include this human-in-the-loop step to ensure that only sensitive correlations that may affect the target task are selected. 
The human-in-the-loop component of our pipeline, like previous efforts~\cite{manjunatha2019explicit, li2021discover, wu2023discover, zhang2023diagnosing}, allows flexibility and transparency for bias mitigation. The human involvement is \textit{optional}, that if given any bias features blindly, our method will generate the mitigation approach as described below.

% % \textbf{Augmenting to balance}
% \paragraph{Human-in-the-loop Bias Cluster Selection}
% \tocite other methods that have HIL bias migitation efforts.
\textbf{Bias mitigation via re-weighting.} We use the feature correlation signals inferred from textual descriptions to mitigate vision model bias. Since adjusting sampling weights to intervene model learning for discounted samples is a frequently used approach ~\cite{wang2020towards, ahn2022mitigating, seo2022unsupervised}, and our goal is \emph{not} to design novel sampling strategies, we use a state-of-the-art sampling method~\cite{qraitem2023bias} based on our analyzed feature presence ($\boldsymbol{t}$). We compute data sampling weights (or probabilities) with the goal to de-correlate the presence of the spurious feature\footnote{We use the term ``spurious features'' to reflect the fact that the primary biases of interest in literature often relate to features such as race/gender/disability/sexuality/etc; however, as we see in Table~\ref{tab:table2}, our method is flexible enough to capture more unexpected types of unwarranted correlation, e.g. ``\texttt{cat}'' and ``\texttt{couch}''.} and the target feature. Specifically, We denote the dataset as $D$, the target feature indicator as $\boldsymbol{t}_{y}$, the spurious feature indicator as $\boldsymbol{t}_{s}$, and the and the conditional probability of $\boldsymbol{t}_{s}$ given $\boldsymbol{t}_{y}$ as $P_D(\boldsymbol{t}_{s}| \boldsymbol{t}_{y})$. The new sampling probability for each element in $D$ ensures that feature $s$ is present in the same probability regardless of whether feature $y$ is present or not:
\begin{equation}\label{eq:sample reweighting}
    P_D'(\boldsymbol{t}_{s}=1| \boldsymbol{t}_{y}=0) = P_{D}'(\boldsymbol{t}_{s}=1 | \boldsymbol{t}_{y}=1)
\end{equation}

% To put it formally, We denote the dataset as $D$, target set as $Y$, obtained target feature indicators as 
% $\boldsymbol{t}_{Y} = \{\boldsymbol{t}_{y} : y \in Y\}$, and sensitive feature indicator as $\boldsymbol{t}_{s}$. 
% Further, we denote the original feature distribution  with respect to the targets as  $P_D(\boldsymbol{t}_{s}|\boldsymbol{t}_{Y})$ and $ P_{D}(\neg \boldsymbol{t}_{s} | \boldsymbol{t}_{Y})$, with $P$ representing probability. The sampling weight $W (\boldsymbol{t}_{s} | \boldsymbol{t}_{Y})$ for images with feature $s$ across the targets is:
% \begin{equation}\label{eq:sample reweighting}
%     P_D(\boldsymbol{t}_{s}| \boldsymbol{t}_{Y})  \cdot W (\boldsymbol{t}_{s} | \boldsymbol{t}_{Y}) = P_{D}(\neg \boldsymbol{t}_{s} | \boldsymbol{t}_{Y})
% \end{equation}
This equation ensures independence between $s$ and $y$ across the dataset~\cite{qraitem2023bias}. 
% In the ideal case where the dataset target feature distribution is already balanced w.r.t the sensitive feature, the sampling weight $W (\boldsymbol{t}_{s} | \boldsymbol{t}_{Y})$ would be equal to 1. 
Additionally, we implement randomized augmentations alongside the sampling weights to enhance data diversity. 

\section{Experiments}
\label{sec:experiments}

%https://docs.google.com/drawings/d/1oLBtGZSUBdSc41wPJSlQynTyXwl_SfQFm2A6nr99Hpc/edit
\begin{figure*}[!htbp]
  \centering
\includegraphics[width=1\linewidth]{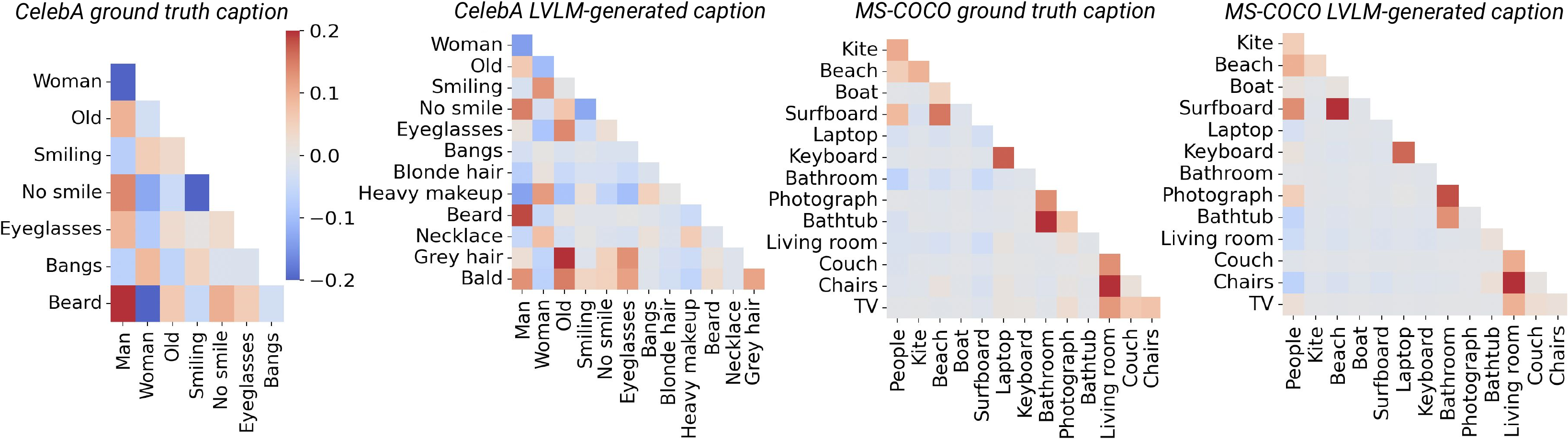}
    \caption{Part of correlated features and their correlation coefficients analyzed by CSBD on CelebA and MS-COCO datasets. Two types of common sense descriptions are used: human-generated (ground truth) and LVLM-generated caption.  
    % Sensitive correlations are circled in yellow, such as ``\texttt{the lady}'' and ``\texttt{smiling}'' with a positive correlation (likely to co-occur). The others are viewed as benign, like ``\texttt{man}'' and ``\texttt{mustache}''.
    }
  \label{fig:correlation}
\end{figure*}

% We conduct downstream experiments to answer: (1) Do the sensitively correlated features reasoned by CSBD align with model bias on popular benchmark datasets? (2) Can prior correlation-robust methods prevent such bias? (3) Is the reasoned bias curable using simple data-wise intervention? 

\subsection{Datasets}
\label{sec:dataset}

\paragraph{CelebA-Dialog}~\cite{jiang2021talk} is a visual-language dataset including captions for images in CelebA~\cite{liu2015deep}, a popular face recognition benchmark dataset containing 40 facial attributes. Each caption describes 5 labels: ``Bangs'', ``Eyeglasses'', ``Beard'', ``Smiling'', and ``Age''. The benefit of captions in natural language is that they also include other common sense information like pronouns and people titles (phrases like ``the man'' or ``the lady''). We identify feature correlations for this dataset accordingly (see Figure~\ref{fig:correlation}). For classification tasks, we select different target labels to recognize and spurious feature labels to evaluate bias results (here ``label'' refers to the ground truth binary labels used only for evaluation). The labels are selected because their corresponding caption-inferred features are discovered by our method to be highly correlated and thus may cause downstream model bias.
% (Gender is expected to be the sensitive attribute)

\paragraph{MS-COCO 2014}~\cite{lin2014microsoft} is a large-scale scene image datasets including annotations for 80 objects, and each image has 5 descriptive captions generated independently~\cite{chen2015microsoft}. We randomly select 1 caption for each image. Based on our feature correlation analysis, we select 9 target and spurious feature pairs for training downstream models and evaluating bias, as listed in Table~\ref{tab:table2}. For example, we train a model for recognizing ``Dog'' and evaluate if the model shows performance disparity for images with/without ``Frisbee''.  We perform binary classification on each target label, following~\cite{wang2023overwriting}. This is because although MS-COCO is more used for multi-label classification, certain labels will need to be treated as auxiliary features in order to reveal their influence on the target thus study bias caused by feature correlations. 
% It should be noted that the target and sensitive label assignment is not limited to what we experiment on.

% \begin{table}[t]
%   \small
%   \begin{center}
%   \setlength{\tabcolsep}{4.2pt}
%     \scalebox{0.8}{
%     \begin{tabular}{*{10}{c}}
%         \toprule
%                   & \texttt{woman} & \texttt{man} & \texttt{kitchen} & \texttt{table} & \texttt{tennis} & \texttt{cat} \\
%         \midrule
%                   LLM-Human & 0. & 0. & 0. & 0. & 0. & 0.    \\
%             Human-Human  &  &  &  &  & NA & NA    \\
%         \midrule \\
%     \end{tabular}}
%     \caption{ }
%     \label{tab:threshold}
%   \end{center}
% \end{table}

\paragraph{LVLM-generated caption.} We also apply our proposed method to image captions generated using large vision language model (LVLM). This data source is considered since human annotated caption is not always available and is especially expensive on large scale datasets. We use the open-source LLaVA-Next 7B parameter model~\cite{liu2024llavanext} to caption images of CelebA dataset with prompt ``\texttt{Please describe face attributes in this image}'', and MS-COCO dataset with prompt ``\texttt{Please describe the scene in this image}''. We evaluate captioning quality of the model on MS-COCO against five ground truth captions. The SPICE~\cite{spice2016} metric on the validation set has an average score of 0.1583, showing a satisfying ability of the LVLM model to capture image features and their relations. Besides, we evaluate the common sense biases modeled from LVLM-generated caption when applying our proposed method, which is in general consistent with the results obtained with ground truth captions (Figure~\ref{fig:correlation}). 
% the feature category of sport, gender, animal, and home items. 
% We emphasis the common sense knowledge in captions over the other captioning model metrics such as consistency. The difference between LLM and human generated captions, in terms of feature distribution, is no bigger than that between two human annotators.  More evaluation on potential bias/error of using LLM is in the supplementary material.
% More evaluations on the robustness of LLM-generated captions are in the supplementary material. 

\subsection{Implementation}
\label{sec:implementation}

For correlation reasoning, the noun chuncks from the description corpus are encoded as vectors in 512 dimensions using the Universal Sentence Encoder. Dimension reduction is performed using PCA before clustering, ensuring that the sum of the PCA components accounts for a high amount of variance ($>$90\%). 
% The embedding vectors are then scaled to unit norm before being clustered using K-means. We set the category cluster number $C$ to 8 for CelebA and 50 for MS-COCO. These numbers are empirically selected to ensure no obvious discrepancies in discovered clusters: that is, no redundant clusters or multiple categories in same clusters. For the second step which generates feature clusters, the upper bound of the within-cluster variance $\sigma_{max}$ is set to 0.15 for CelebA and 0.5 for MS-COCO; see Sec.~\ref{sec:ablation} for sensitivity analysis. 
The distance criterion for generating feature clusters is set to $z=1.0$; see ablation sections for sensitivity analysis. We use Chi-square test~\cite{Pearson1900} to verify significance of derived correlation coefficients. 
% Empirically, feature correlations higher than $z$ cause bias on the datasets we used. 

% all pair-wise correlations regardless of magnitude will be recorded. (Supplementary)

For the downstream training, we use the same training, validation, and testing split as in \cite{zhang2022correct} for CelebA dataset, and as in \cite{misra2016seeing} for MS-COCO dataset. 
% zee: todo
However, because of label scarcity of individual objects relative to the entire dataset  (for example, only 3\% of images in MS-COCO contain ``Cat'' and 2\% of images contain ``Kite''), to avoid class imbalance problem introducing confounding bias to the classifier, we randomly take subsets of the dataset splits which include 50\% of each target label. Experiments are based on three independent runs which provide the mean and standard deviation results.

\begin{table*}[!htbp]
  \small
  \begin{center}
    \label{tab:table1}
    \scalebox{0.92}{
    \begin{tabular}{*{14}{c}}
    
        \toprule
                  & & \multicolumn{2}{c}{No mitigation} & 
                  \multicolumn{2}{c}{LfF} & \multicolumn{2}{c}{DebiAN} &
                  \multicolumn{2}{c}{PGD} &
                  \multicolumn{2}{c}
                  {BPA} &
                  \multicolumn{2}{c}{CSBD (ours)} \\
                  \cmidrule(lr){3-4}
                  \cmidrule(lr){5-6}
                  \cmidrule(lr){7-8}
                  \cmidrule(lr){9-10}
                  \cmidrule(lr){11-12}
                  \cmidrule(lr){13-14}
                   Target & Spurious &  Wst. ($\uparrow$)& Avg. ($\uparrow$)& Wst. & Avg.  & Wst.  & Avg. & Wst. & Avg.  &
                  Wst.  & Avg.  &
                  Wst.  & Avg. \\
        \midrule
        \midrule
                   Smiling  & Male & 0.894  & \textbf{0.919}  & 0.730  & 0.841  & 0.895  & 0.916  & 0.888   & 0.917 & \underline{0.899}  & 0.912  & \textbf{0.908}  & \underline{0.918}     \\
           Eyeglasses   & Male & 0.962  & \underline{0.986}   &  0.912   & 0.963 & 0.967  & 0.979  & 0.966  & 0.984    & \underline{0.971}  & 0.983 &  \textbf{0.973}  & \textbf{0.987}   \\ 
               Young *  & Eyeglasses * & \underline{0.652}   &  \underline{0.818}  &  0.640  &  0.805  &  0.609  & 0.796 & 0.646 & 0.817  & 0.638 &  0.806 & \textbf{0.655} & \textbf{0.820}  \\ 
           Receding hairline *  & Grey hair *  & 0.186  &  0.650  &  0.174  &  0.648  & 0.198 & \textbf{0.680} & 0.225 & 0.672 & \textbf{0.351} & 0.521 & \underline{0.337} & \textbf{0.680} \\ 
        \bottomrule

    \end{tabular}}
    \caption{Treating common sense bias towards the target label (Target) among the spurious feature label (Spurious) on CelebA. The starred labels (*) represent that CSBD uses LVLM-generated descriptions to discover and mitigate bias. The metric for debiasing effectiveness: the worst performing group (Wst.), and average accuracy of all target-spurious feature groups (Avg.), are reported. The best results are in bold and the second best results are underlined.}
    \label{tab:table1}
  \end{center}
\end{table*}

\begin{table*}[!htbp]
  \small
  \begin{center}
    \scalebox{0.94}{
    \begin{tabular}{*{14}{c}}
       \toprule
                  & & \multicolumn{2}{c}{No mitigation} & 
                  \multicolumn{2}{c}{LfF} & \multicolumn{2}
                  {c}{DebiAN} & \multicolumn{2}{c}{PGD} &
                  \multicolumn{2}{c}
                  {BPA} &
                  \multicolumn{2}{c}{CSBD (ours)} \\
                  \cmidrule(lr){3-4}
                  \cmidrule(lr){5-6}
                  \cmidrule(lr){7-8}
                  \cmidrule(lr){9-10}
                  \cmidrule(lr){11-12}
                  \cmidrule(lr){13-14}
                   Target & Spurious & Wst.($\uparrow$) & Avg. ($\uparrow$) & Wst.  & Avg. & Wst.  & Avg. & Wst.  & Avg.  & Wst.  & Avg.  
                   & Wst. & Avg.  \\
        \midrule
        \midrule
             Cat   & Couch & 0.849  & 0.935  & 0.800  & 0.916  & 0.821  & 0.902  &  \underline{0.884}  & \underline{0.940}  & 0.853  & 0.925   & \textbf{0.906}  & \textbf{0.941}  \\
              Cat   & Keyboard &  \underline{0.948}  & 0.972  & 0.870  & 0.938  & 0.917  & 0.949  & 0.947  & \textbf{0.977}  & 0.890  & 0.952 & \textbf{0.953}  & \underline{0.976}  \\
             Dog  &  Frisbee & \underline{0.883}  & 0.927  & 0.821  & 0.893  & 0.845  & 0.895  & 0.872   & \textbf{0.940}  & 0.854  & 0.905  & \textbf{0.903}  & \underline{0.934}  \\
           TV  & Chair & 0.726  & 0.893  & 0.690  & 0.867  & 0.718  & 0.876  & \underline{0.821}   &  \underline{0.907}  & 0.758   &  0.887  &  \textbf{0.842}  & \textbf{0.911}  \\
            Umbrella & Person & 0.761  & 0.880  & 0.771  & 0.854  & 0.716  & 0.857  &  0.763   & \underline{0.887}  &  \underline{0.777}  & 0.878   & \textbf{0.814}  & \textbf{0.892} \\
            Kite & Person & 0.937  & \textbf{0.963}  & 0.854  & 0.923   & \underline{0.941}  & \underline{0.960}  &   0.917  & 0.955  & 0.921  & 0.957   & \textbf{0.945}  & \underline{0.960}   \\
                Wine glasses * & Person *  & \underline{0.848}   & 0.888   &  0.834  &  0.869  & 0.838  & 0.873 & 0.828 & 0.869 & 0.844 & \underline{0.890} & \textbf{0.860} & \textbf{0.894} \\
            Pizza * & Oven * & 0.828   & 0.916   & 0.820 &  0.861   & 0.827 & 0.910 & 0.828 & 0.903 & \underline{0.854} & \underline{0.925} & \textbf{0.862} & \textbf{0.932} \\
        \bottomrule
    \end{tabular}}
    \caption{Treating part of the discovered biases for binary object classification task on MS-COCO dataset.  }
    \label{tab:table2}
  \end{center}
\end{table*}

\subsection{Results}
\label{sec:results}

Bias discovery and mitigation are both needed for reliable model training.  While the emphasis of the proposed method is to analyze bias from datasets, not primarily on mitigation algorithm, it is important to verify on downstream tasks that the method ``discovers'' correlations in the dataset that the model actually takes a shortcut for learning. So in this section, we first present bias discovery results on the two benchmark datasets. Then, we analyze robustness of the correlation derivation, with respect to potential errors and incompleteness of descriptions. Finally, we perform thorough downstream evaluations to show that reducing feature dependence analyzed by the proposed method serves as a practical and effective debiasing objective. 

% with respect to the inconsistency between common sense feature distribution and label class distribution,

% Similarly, verifying that the method ``discovers'' correlations in the dataset that the model actually relies on is demonstrated by the improved downstream results after reducing feature dependence.

\subsubsection{Correlation discovery results}
\label{sec:corr_results}

Figure~\ref{fig:correlation} presents selected correlated feature pairs for CelebA and MS-COCO dataset. It shows that both human-annotated and LVLM-generated captions contain rich common sense information that can be distilled into distinctive features. The correlation between the obtained features are consistent across different image descriptions, highlighting the description-agnostic and generalization strength of the proposed method. We note that Figure~\ref{fig:correlation} shows limited CelebA feature correlations with the ground truth caption because the caption is designed to annotate only five features.

The description-derived features include those not represented in the dataset labels and reveal highly correlated ones, such as ``\texttt{city street}'' and ``\texttt{bus}''. The spuriously correlated features are frequently seen but not generalizable to all samples, so models cannot predict one feature robustly if relying on the other: for example, recognition of a bus in an image might depend not on its appearance but on whether the background is a city street. On CelebA dataset, facial features ``\texttt{smiling}'',  ``\texttt{eyeglasses}'', ``\texttt{bangs}'', etc, are spuriously correlated to gender features ``\texttt{man}'', ``\texttt{woman}'', and age features ``\texttt{young}'', etc. Other strong correlations include ``\texttt{Receding hairline}'' and ``\texttt{grey hair}'' which aligns with common sense but lacks robustness. We designate naturally related features like ``\texttt{mouth}'' and ``\texttt{smiling}'' as benign. On MS-COCO dataset, benign correlations include ``\texttt{frisbee}'' and ``\texttt{a game}'', ``\texttt{waves}'' and ``\texttt{ocean}'', etc. The spurious ones include ``\texttt{cat}'' and ``\texttt{couch}'', ``\texttt{kite}'' and ``\texttt{people}'', ``\texttt{bathroom}'' and ``\texttt{photograph}'', etc. The full list of correlated feature pairs and their coefficient values are in the Supplementary.

% emantic parsing for descriptions generates 
% $M=253,\!467$ phrases. CSBD outputs 220 correlated feature pairs for the MS-COCO dataset.

% semantic parsing of the image descriptions generates $M=1,\!283,\!552$ phrases for CelebA dataset, and applying CSBD outputs 72 feature pairs that have an absolute correlation coefficient higher than $z=0.05$. By manual check, we determine that the features ``\texttt{smiling}'' and ``\texttt{eyeglasses}'' are both sensitively correlated to people-title features ``\texttt{the man, the guy, ...}'' and ``\texttt{the lady, the girl, ...}'', and we designate other naturally related features like ``\texttt{mouth}'' and ``\texttt{smiling}'' as benign. 

% The full list of correlated feature pairs for both dataset, as one of the main outcomes of our analysis, are attached with the Supplementary. The examples with image illustrations for both datasets are in \cref{fig:correlation}. We additionally mark part of sensitive correlations (intuitive but not robust) in \cref{tab:coco_corr1}; the description-derived features include those not represented in the dataset labels, and expose highly correlated ones, such as ``\texttt{bathroom}'' and ``\texttt{vase}''.

% % The correlation values between sensitive features, along with selected benign correlations, are visualized in Figure~\ref{fig:correlation} (Left). 

% % The discovered correlations quantify feature distribution imbalance existing in a dataset, which may lead to model bias for the related training tasks. 

\subsubsection{Correlation robustness evaluation}
% Robustness to caption quality

First, we test the robustness of analyzed feature presence, considering that unlike dataset label, description annotators may only select features relevant or noticeable to them, causing incompleteness or faults. Indeed, prior studies ~\cite{yu2022automated, wang2022measuring} have highlighted the incorrect and low-quality captions in MS-COCO. We randomly select 20 common sense features derived from MS-COCO captions and compare them with their corresponding labels. We compute the Pearson correlation coefficient between the presence of each feature and label across the dataset, obtaining a coefficient of 0.545 (95\% confidence interval 0.153 to 0.937). This indicates a moderate correlation between label and feature distribution, which varies across different features. The inconsistency between common sense feature and label distribution is expected, influenced by both caption quality (captions may incorrectly describe ground truth label) and text embedding clustering (text with similar semantics may not be encoded closely in the representation space). 

Next, we test the robustness of discovered feature correlations (or bias, to avoid confusion with the correlation test here, we use the term ``bias'' ) under this circumstance. Bias existence between each feature and each of the remaining features is computed. The same bias measurement is then performed using labels. The correlation coefficient between the two measurements is 0.781 (95\% confidence interval 0.417 to 1.0), which is much higher and shows lower variance than the result for feature presence. The analysis indicates that the proposed method operates not at the level of individual images but rather at the dataset level to identify trends in correlations, thus showing certain robustness to noise and incompleteness present in image descriptions.

% First, consistency of derived feature distribution $\boldsymbol{t}_{f}$ to ground truth label distribution is evaluated. This consistency is decided by both caption quality (whether captions incorrectly describe features) and text embedding clustering (whether text with similar semantics is clustered together using our method). Second, we compare the correlation derived from features and that from labels. We randomly select 20 features and manually aligned labels of MS-COCO to show the results:

\begin{figure*}[!htbp]
  \centering
    \includegraphics[width=1\linewidth]{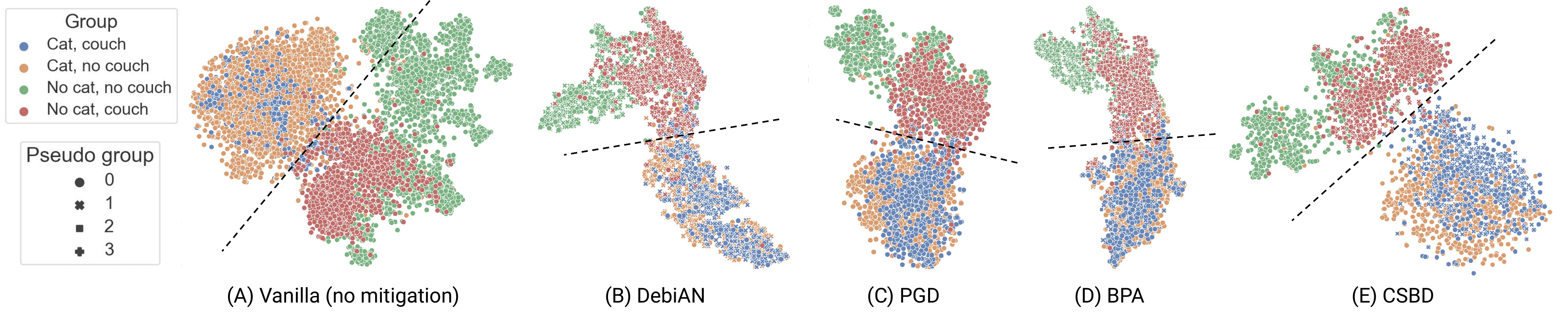}
    \caption{T-SNE results for sample embeddings of cat classifiers (``Couch'' as sensitive label) trained with MS-COCO. Target label: ``Cat'' (blue \& orange) and ``No cat'' (green \& red) are most linearly separable (model decision boundary is indicated by the dotted straight line) with CSBD compared to other baselines. For bias discovery, methods are marked with their assigned bias-aligning and bias-conflict samples (Pseudo group). Only CSBD discovers thus treats the sensitive feature couch (E).
    % Methods which split bias-aligning and bias-conflict samples (Pseudo group) are marked with their group assignment by $\bullet$ and $\times$. Comparing to real bias (``Couch'' \textcolor{red}{$\bullet$} \textcolor{blue}{$\bullet$} and ``No couch'' \textcolor{green}{$\bullet$} \textcolor{orange}{$\bullet$}).
    }
  \label{fig:embedding}
\end{figure*}

\subsubsection{Bias mitigation results}
\label{sec:bias_results}
Spurious feature correlations discovered by CSBD can indicate and help mitigate downstream model bias, as shown by image classification results on CelebA and MS-COCO dataset in Table~\ref{tab:table1} and Table~\ref{tab:table2}.  We use the worst group performance (Wst.) to evaluate model bias, and average group performance (Avg.) to evaluate model quality, following~\cite{zhang2022correct, ahn2022mitigating, seo2022unsupervised}. The groups here are all combinations of target label and spurious feature label pairs as in~\cite{zhang2022correct}. The comparison baselines are unsupervised bias mitigation methods. They include LfF~\cite{nam2020learning} which upweights failure samples, DebiAN~\cite{li2022discover} which uses additional bias predictor, and re-sampling methods PGD~\cite{ahn2022mitigating} based on sample gradients, and BPA based on latent space clusters~\cite{seo2022unsupervised}. 

Baseline model without bias mitigation presents a gap between Wst. and Avg. for all tasks, indicating model biases that align with our reasoning on the datasets. For example, images containing chair but no TV obtain an accuracy of 0.726 for TV classification, 19\% lower than the average accuracy. Face images to recognize receding hairline with dark hair show an accuracy of 0.186, while the recognition with grey hair has an accuracy over 0.99 because of strong correlation between these two features. 
% % The ``\texttt{tv screen}'' and ``\texttt{chair}'' correlation discovered by our method tells how the sensitive feature affects downstream model decision rule. 
CSBD successfully mitigates model bias (most improved Wst.) on all targets, outperforming the baseline with no mitigation and other de-biasing methods. Importantly, model classification quality is not degraded by using our method: CSBD always obtains the highest or the second Avg. Bias mitigation based on LVLM generated descriptions are effective on both datasets, which shows our method's generalizability to different description corpus and wide applicability to bias mitigation problems.

% % Differentiate CSBD to feature-based methods: The bias-guiding samples prior approaches identify are noisy; Feature clustering approaches might cluster attributes that have very similar features but very different concepts.

% % common-sense feature clusters are  used for sensitive correlation discovery and to re-arrange training sample accordingly. The bias groups in evaluation are split according to dataset ground truth label.

\section{Discussion}
\label{sec:discussion}

We investigate the advantage of our method for bias mitigation by checking how well they separate target objects not affected by spurious features in latent space. We visualize image embeddings extracted from the last fully connected layer of the ``Cat'' classifier trained with MS-COCO. Model bias can be observed from the baseline (no mitigation) (Figure~\ref{fig:embedding} (A)): while the ``Cat'' targets are separable for samples without the spurious feature ``Couch'', for images \emph{with} the spurious feature, the model prediction boundary for the targets is blurred (blue and red dots are partially mixed).  

The same embedding analysis is performed for bias mitigation methods\footnote{Here only showing methods outperforming no mitigation.}. As shown in Figure~\ref{fig:embedding} (B-D), DebiAN, PGD, and BPA do not fix model's biased decision rule, that the model cannot distinguish two targets (``Cat'' and `No cat'' samples), indicated by the dotted straight line. Instead, the methods only mix samples within a target class more closely. The reason is that the methods identify discrepancy between sample representations or gradients as bias and try to mitigate the discrepancy. However, the detected discrepancy fails to accurately capture the  unwanted feature responsible for mis-classifying bias-conflicting samples (those near the decision boundary), so the black-box way of mitigation proves less effective for de-biasing.

\begin{table}[ht!]
  \small
  \begin{center}
    \scalebox{0.9}{
    \begin{tabular}{*{12}{c}}
        \toprule
                 $z$ & 0.5 & 0.8 & 0.9 & 1 & 1.2 & 1.5  \\
        \midrule
        \midrule
                  $\phi$  & 0.0608 & 0.0608 & 0.0908 & 0.0908 & 0.107 & 0.175   \\
            Wst. ($\uparrow$)  & 0.971 & 0.972  & 0.973  & 0.973   & 0.973 & 0.972   \\
                  Avg. ($\uparrow$) & 0.986  & 0.986 & 0.986  & 0.987   & 0.987 & 0.986  \\  
        \midrule \\
    \end{tabular}}
    \caption{Sensitivity analysis on the distance threshold $z$ used for splitting feature clusters. $z$ within a reasonable range can  discover consistent correlated features ($\phi>0$) and produce similar bias mitigation results (Wst.), between ``\texttt{eyeglasses}" and ``\texttt{man}" features on CelebA. 
    % $z$ of 0.01 and 0.05 generate redundant clusters that impact correlation measurement thus computed intervention. $z$ bigger than 0.2 generate clusters with ambiguous semantic meanings so no correlation is discovered (NA). 
    }
    \label{tab:threshold}
  \end{center}
\end{table}

% Show different Z in Supplementary
\textbf{Ablation on the clustering threshold} Feature clusters are obtained by agglomerative clustering on noun chunk embeddings so that the complete linkage between two clusters is lower than a hyper-parameter $z$. This tune-able parameter affects the granularity of discovered feature clusters and whether reasonable bias can be derived. As shown in Table~\ref{tab:threshold}, the bias mitigation (Wst.) and classification (Avg.) results are robust to different $z$, which indicates that feature clusters and their distributions are captured precisely in all cases. The granularity and content of feature clusters change with different $z$, which result in different correlation coefficient $\phi$. Here, ``\texttt{eyeglasses}'' and ``\texttt{sunglasses}'' are one feature with $z=\{1.2, 1.5\}$ and shows a stronger correlation to ``\texttt{man}'' feature, but they are separate features with smaller $z$ and ``\texttt{eyeglasses}'' alone shows weaker correlation to ``\texttt{man}''. The selection of $z$ should be tuned based on the granularity of features that practitioners hope to derive and find biases with. For example, if $z$ is too large (we test with $z=5$), gender clusters will be replaced by a ``\texttt{people}'' cluster, thus the bias:  man wearing glasses more will not be discovered on the CelebA dataset.   

\textbf{Ablation on re-sampling weights.} We ablate the bias mitigation step by skewing the  sampling weights $W$ analyzed by CSBD. Specifically, for any images that need to be over-sampled ($W[i] = 1 + p$ for the $i^{th}$ sample), we first double their increased weight by $W[i] = 1 + 2*p$. The same downstream task to classify ``Cat" with spurious attribute ``Couch" on MS-COCO is performed. The model obtains \underline{Wst.: 0.904, Avg.: 0.940}. Then, we only use half of the increased weight by $W[i] = 1 + 0.5*p$, and the model obtains \underline{Wst.: 0.881, Avg.: 0.936}. Compared to the results in Table~\ref{tab:table2}, though training with skewed weights still outperforms the vanilla setting with no intervention, the bias mitigation and accuracy results are slightly discounted. It shows that the data re-sampling weights computed from CSBD is the key for effective bias mitigation.

\section{Conclusion and Limitation}
\label{sec:conclusion}

This work proposes to use textural descriptions of a dataset for systematic bias discovery and mitigation. The approach leads to a new perspective on bias problems regarding how we can use knowledge in natural language to precisely reduce a concerned feature's influence to image classification tasks. Our algorithm applies common sense reasoning to descriptions, which cover a wide range of image features not annotated by the label set, and indicates how feature correlations to the target will cause downstream bias. Experiment results show novel spurious correlations and model biases discovered for two  vision benchmark datasets, and the state-of-the-art bias mitigation based on the discovery.

% to add: reference to Bert embedding

\paragraph{Limitations and future work.} Though our method can model correlations between features beyond the labeled cohort, the discovered biases are limited to features included in textual descriptions. Besides, there are multiple topics to continue in this direction: First, we only study pairwise feature correlations, but higher degrees of correlations and how they affect model performance could be explored. Second, our method has not yet been tested in other vision recognition tasks and models, such as object detection and segmentation. However, since the method for discovering spurious feature correlations is agnostic to downstream tasks, this can be a direct next step.

% The description quality is not required as high as labels and will not break the method. For example, the description for an image with cat and couch could be ``\texttt{The cats are laying together looking very comfortable.}'' which misses the couch feature.

% Future Work: (1) Hybrid image and text feature 

% If text descriptions cover other types of attributes, we could compete. 

{
    \small
    \bibliographystyle{ieeenat_fullname}
    \bibliography{main}

\begin{thebibliography}{54}
\providecommand{\natexlab}[1]{#1}
\providecommand{\url}[1]{\texttt{#1}}
\expandafter\ifx\csname urlstyle\endcsname\relax
  \providecommand{\doi}[1]{doi: #1}\else
  \providecommand{\doi}{doi: \begingroup \urlstyle{rm}\Url}\fi

\bibitem[Ahn et~al.(2023)Ahn, Kim, and Yun]{ahn2022mitigating}
Sumyeong Ahn, Seongyoon Kim, and Se-young Yun.
\newblock Mitigating dataset bias by using per-sample gradient.
\newblock In \emph{Eleventh International Conference on Learning Representations}. ICLR, 2023.

\bibitem[Anderson et~al.(2016)Anderson, Fernando, Johnson, and Gould]{spice2016}
Peter Anderson, Basura Fernando, Mark Johnson, and Stephen Gould.
\newblock Spice: Semantic propositional image caption evaluation.
\newblock In \emph{ECCV}, 2016.

\bibitem[Antol et~al.(2015)Antol, Agrawal, Lu, Mitchell, Batra, Zitnick, and Parikh]{antol2015vqa}
Stanislaw Antol, Aishwarya Agrawal, Jiasen Lu, Margaret Mitchell, Dhruv Batra, C~Lawrence Zitnick, and Devi Parikh.
\newblock Vqa: Visual question answering.
\newblock In \emph{Proceedings of the IEEE international conference on computer vision}, pages 2425--2433, 2015.

\bibitem[Basu et~al.(2023)Basu, Babu, and Pruthi]{basu2023inspecting}
Abhipsa Basu, R~Venkatesh Babu, and Danish Pruthi.
\newblock Inspecting the geographical representativeness of images from text-to-image models.
\newblock \emph{arXiv preprint arXiv:2305.11080}, 2023.

\bibitem[Brown et~al.(2023)Brown, Tomasev, Freyberg, Liu, Karthikesalingam, and Schrouff]{brown2023detecting}
Alexander Brown, Nenad Tomasev, Jan Freyberg, Yuan Liu, Alan Karthikesalingam, and Jessica Schrouff.
\newblock Detecting shortcut learning for fair medical ai using shortcut testing.
\newblock \emph{Nature Communications}, 14\penalty0 (1):\penalty0 4314, 2023.

\bibitem[Cer et~al.(2018)Cer, Yang, Kong, Hua, Limtiaco, John, Constant, Guajardo-Cespedes, Yuan, Tar, et~al.]{cer2018universal}
Daniel Cer, Yinfei Yang, Sheng-yi Kong, Nan Hua, Nicole Limtiaco, Rhomni~St John, Noah Constant, Mario Guajardo-Cespedes, Steve Yuan, Chris Tar, et~al.
\newblock Universal sentence encoder.
\newblock \emph{arXiv preprint arXiv:1803.11175}, 2018.

\bibitem[Chen et~al.(2015)Chen, Fang, Lin, Vedantam, Gupta, Doll{\'a}r, and Zitnick]{chen2015microsoft}
Xinlei Chen, Hao Fang, Tsung-Yi Lin, Ramakrishna Vedantam, Saurabh Gupta, Piotr Doll{\'a}r, and C~Lawrence Zitnick.
\newblock Microsoft coco captions: Data collection and evaluation server.
\newblock \emph{arXiv preprint arXiv:1504.00325}, 2015.

\bibitem[Cram{\'e}r(1946)]{Cramr1946MathematicalMO}
Harald Cram{\'e}r.
\newblock Mathematical methods of statistics.
\newblock 1946.

\bibitem[Diomataris et~al.(2021)Diomataris, Gkanatsios, Pitsikalis, and Maragos]{diomataris2021grounding}
Markos Diomataris, Nikolaos Gkanatsios, Vassilis Pitsikalis, and Petros Maragos.
\newblock Grounding consistency: Distilling spatial common sense for precise visual relationship detection.
\newblock In \emph{Proceedings of the IEEE/CVF International Conference on Computer Vision}, pages 15911--15920, 2021.

\bibitem[He et~al.(2016)He, Zhang, Ren, and Sun]{he2016deep}
Kaiming He, Xiangyu Zhang, Shaoqing Ren, and Jian Sun.
\newblock Deep residual learning for image recognition.
\newblock In \emph{Proceedings of the IEEE conference on computer vision and pattern recognition}, pages 770--778, 2016.

\bibitem[Honnibal and Montani(2017)]{honnibal2017spacy}
Matthew Honnibal and Ines Montani.
\newblock spacy 2: Natural language understanding with bloom embeddings, convolutional neural networks and incremental parsing.
\newblock \emph{To appear}, 7\penalty0 (1):\penalty0 411--420, 2017.

\bibitem[Jiang et~al.(2021)Jiang, Huang, Pan, Loy, and Liu]{jiang2021talk}
Yuming Jiang, Ziqi Huang, Xingang Pan, Chen~Change Loy, and Ziwei Liu.
\newblock Talk-to-edit: Fine-grained facial editing via dialog.
\newblock In \emph{Proceedings of the IEEE/CVF International Conference on Computer Vision}, pages 13799--13808, 2021.

\bibitem[Kim et~al.(2021)Kim, Lee, and Choo]{kim2021biaswap}
Eungyeup Kim, Jihyeon Lee, and Jaegul Choo.
\newblock Biaswap: Removing dataset bias with bias-tailored swapping augmentation.
\newblock In \emph{Proceedings of the IEEE/CVF International Conference on Computer Vision}, pages 14992--15001, 2021.

\bibitem[Krishnakumar et~al.(2021)Krishnakumar, Prabhu, Sudhakar, and Hoffman]{krishnakumar2021udis}
Arvindkumar Krishnakumar, Viraj Prabhu, Sruthi Sudhakar, and Judy Hoffman.
\newblock Udis: Unsupervised discovery of bias in deep visual recognition models.
\newblock In \emph{British Machine Vision Conference (BMVC)}, page~3, 2021.

\bibitem[Lake et~al.(2017)Lake, Ullman, Tenenbaum, and Gershman]{lake2017building}
Brenden~M Lake, Tomer~D Ullman, Joshua~B Tenenbaum, and Samuel~J Gershman.
\newblock Building machines that learn and think like people.
\newblock \emph{Behavioral and brain sciences}, 40:\penalty0 e253, 2017.

\bibitem[Lee et~al.(2023)Lee, Park, Kim, Lee, Choi, and Choo]{lee2023revisiting}
Jungsoo Lee, Jeonghoon Park, Daeyoung Kim, Juyoung Lee, Edward Choi, and Jaegul Choo.
\newblock Revisiting the importance of amplifying bias for debiasing.
\newblock In \emph{Proceedings of the AAAI Conference on Artificial Intelligence}, pages 14974--14981, 2023.

\bibitem[Li et~al.(2023)Li, Vo, and Nakayama]{li2023partition}
Jiaxuan Li, Duc~Minh Vo, and Hideki Nakayama.
\newblock Partition-and-debias: Agnostic biases mitigation via a mixture of biases-specific experts.
\newblock In \emph{Proceedings of the IEEE/CVF International Conference on Computer Vision}, pages 4924--4934, 2023.

\bibitem[Li and Xu(2021)]{li2021discover}
Zhiheng Li and Chenliang Xu.
\newblock Discover the unknown biased attribute of an image classifier.
\newblock In \emph{Proceedings of the IEEE/CVF International Conference on Computer Vision}, pages 14970--14979, 2021.

\bibitem[Li et~al.(2022)Li, Hoogs, and Xu]{li2022discover}
Zhiheng Li, Anthony Hoogs, and Chenliang Xu.
\newblock Discover and mitigate unknown biases with debiasing alternate networks.
\newblock In \emph{European Conference on Computer Vision}, pages 270--288. Springer, 2022.

\bibitem[Lim et~al.(2023)Lim, Kim, Kim, Ahn, Shin, Yang, and Han]{lim2023biasadv}
Jongin Lim, Youngdong Kim, Byungjai Kim, Chanho Ahn, Jinwoo Shin, Eunho Yang, and Seungju Han.
\newblock Biasadv: Bias-adversarial augmentation for model debiasing.
\newblock In \emph{Proceedings of the IEEE/CVF conference on computer vision and pattern recognition}, pages 3832--3841, 2023.

\bibitem[Lin et~al.(2014)Lin, Maire, Belongie, Hays, Perona, Ramanan, Doll{\'a}r, and Zitnick]{lin2014microsoft}
Tsung-Yi Lin, Michael Maire, Serge Belongie, James Hays, Pietro Perona, Deva Ramanan, Piotr Doll{\'a}r, and C~Lawrence Zitnick.
\newblock Microsoft coco: Common objects in context.
\newblock In \emph{Computer Vision--ECCV 2014: 13th European Conference, Zurich, Switzerland, September 6-12, 2014, Proceedings, Part V 13}, pages 740--755. Springer, 2014.

\bibitem[Liu et~al.(2021)Liu, Haghgoo, Chen, Raghunathan, Koh, Sagawa, Liang, and Finn]{liu2021just}
Evan~Z Liu, Behzad Haghgoo, Annie~S Chen, Aditi Raghunathan, Pang~Wei Koh, Shiori Sagawa, Percy Liang, and Chelsea Finn.
\newblock Just train twice: Improving group robustness without training group information.
\newblock In \emph{International Conference on Machine Learning}, pages 6781--6792. PMLR, 2021.

\bibitem[Liu et~al.(2024)Liu, Li, Li, Li, Zhang, Shen, and Lee]{liu2024llavanext}
Haotian Liu, Chunyuan Li, Yuheng Li, Bo Li, Yuanhan Zhang, Sheng Shen, and Yong~Jae Lee.
\newblock Llava-next: Improved reasoning, ocr, and world knowledge, 2024.

\bibitem[Liu et~al.(2015)Liu, Luo, Wang, and Tang]{liu2015deep}
Ziwei Liu, Ping Luo, Xiaogang Wang, and Xiaoou Tang.
\newblock Deep learning face attributes in the wild.
\newblock In \emph{Proceedings of the IEEE international conference on computer vision}, pages 3730--3738, 2015.

\bibitem[Manjunatha et~al.(2019)Manjunatha, Saini, and Davis]{manjunatha2019explicit}
Varun Manjunatha, Nirat Saini, and Larry~S Davis.
\newblock Explicit bias discovery in visual question answering models.
\newblock In \emph{Proceedings of the IEEE/CVF Conference on Computer Vision and Pattern Recognition}, pages 9562--9571, 2019.

\bibitem[Matthews(1975)]{Matthews1975ComparisonOT}
Brian~W. Matthews.
\newblock Comparison of the predicted and observed secondary structure of t4 phage lysozyme.
\newblock \emph{Biochimica et biophysica acta}, 405 2:\penalty0 442--51, 1975.

\bibitem[Misra et~al.(2016)Misra, Lawrence~Zitnick, Mitchell, and Girshick]{misra2016seeing}
Ishan Misra, C Lawrence~Zitnick, Margaret Mitchell, and Ross Girshick.
\newblock Seeing through the human reporting bias: Visual classifiers from noisy human-centric labels.
\newblock In \emph{Proceedings of the IEEE conference on computer vision and pattern recognition}, pages 2930--2939, 2016.

\bibitem[Nam et~al.(2020)Nam, Cha, Ahn, Lee, and Shin]{nam2020learning}
Junhyun Nam, Hyuntak Cha, Sungsoo Ahn, Jaeho Lee, and Jinwoo Shin.
\newblock Learning from failure: De-biasing classifier from biased classifier.
\newblock \emph{Advances in Neural Information Processing Systems}, 33:\penalty0 20673--20684, 2020.

\bibitem[Park et~al.(2022)Park, Lee, Lee, Hwang, Kim, and Byun]{park2022fair}
Sungho Park, Jewook Lee, Pilhyeon Lee, Sunhee Hwang, Dohyung Kim, and Hyeran Byun.
\newblock Fair contrastive learning for facial attribute classification.
\newblock In \emph{Proceedings of the IEEE/CVF Conference on Computer Vision and Pattern Recognition}, pages 10389--10398, 2022.

\bibitem[Pearson(1900)]{Pearson1900}
Karl Pearson.
\newblock X. on the criterion that a given system of deviations from the probable in the case of a correlated system of variables is such that it can be reasonably supposed to have arisen from random sampling.
\newblock \emph{The London, Edinburgh, and Dublin Philosophical Magazine and Journal of Science}, 50\penalty0 (302):\penalty0 157--175, 1900.

\bibitem[Qraitem et~al.(2023)Qraitem, Saenko, and Plummer]{qraitem2023bias}
Maan Qraitem, Kate Saenko, and Bryan~A Plummer.
\newblock Bias mimicking: A simple sampling approach for bias mitigation.
\newblock In \emph{Proceedings of the IEEE/CVF Conference on Computer Vision and Pattern Recognition}, pages 20311--20320, 2023.

\bibitem[Radford et~al.(2021)Radford, Kim, Hallacy, Ramesh, Goh, Agarwal, Sastry, Askell, Mishkin, Clark, et~al.]{radford2021learning}
Alec Radford, Jong~Wook Kim, Chris Hallacy, Aditya Ramesh, Gabriel Goh, Sandhini Agarwal, Girish Sastry, Amanda Askell, Pamela Mishkin, Jack Clark, et~al.
\newblock Learning transferable visual models from natural language supervision.
\newblock In \emph{International conference on machine learning}, pages 8748--8763. PMLR, 2021.

\bibitem[Ramaswamy et~al.(2021)Ramaswamy, Kim, and Russakovsky]{ramaswamy2021fair}
Vikram~V Ramaswamy, Sunnie~SY Kim, and Olga Russakovsky.
\newblock Fair attribute classification through latent space de-biasing.
\newblock In \emph{Proceedings of the IEEE/CVF conference on computer vision and pattern recognition}, pages 9301--9310, 2021.

\bibitem[Rombach et~al.(2022)Rombach, Blattmann, Lorenz, Esser, and Ommer]{rombach2022high}
Robin Rombach, Andreas Blattmann, Dominik Lorenz, Patrick Esser, and Bj{\"o}rn Ommer.
\newblock High-resolution image synthesis with latent diffusion models.
\newblock In \emph{Proceedings of the IEEE/CVF conference on computer vision and pattern recognition}, pages 10684--10695, 2022.

\bibitem[Sagawa et~al.(2019)Sagawa, Koh, Hashimoto, and Liang]{sagawa2019distributionally}
Shiori Sagawa, Pang~Wei Koh, Tatsunori~B Hashimoto, and Percy Liang.
\newblock Distributionally robust neural networks for group shifts: On the importance of regularization for worst-case generalization.
\newblock \emph{arXiv preprint arXiv:1911.08731}, 2019.

\bibitem[Seo et~al.(2022)Seo, Lee, and Han]{seo2022unsupervised}
Seonguk Seo, Joon-Young Lee, and Bohyung Han.
\newblock Unsupervised learning of debiased representations with pseudo-attributes.
\newblock In \emph{Proceedings of the IEEE/CVF Conference on Computer Vision and Pattern Recognition}, pages 16742--16751, 2022.

\bibitem[Singh et~al.(2020)Singh, Mahajan, Grauman, Lee, Feiszli, and Ghadiyaram]{singh2020don}
Krishna~Kumar Singh, Dhruv Mahajan, Kristen Grauman, Yong~Jae Lee, Matt Feiszli, and Deepti Ghadiyaram.
\newblock Don't judge an object by its context: Learning to overcome contextual bias.
\newblock In \emph{Proceedings of the IEEE/CVF Conference on Computer Vision and Pattern Recognition}, pages 11070--11078, 2020.

\bibitem[Sohoni et~al.(2020)Sohoni, Dunnmon, Angus, Gu, and R{\'e}]{sohoni2020no}
Nimit Sohoni, Jared Dunnmon, Geoffrey Angus, Albert Gu, and Christopher R{\'e}.
\newblock No subclass left behind: Fine-grained robustness in coarse-grained classification problems.
\newblock \emph{Advances in Neural Information Processing Systems}, 33:\penalty0 19339--19352, 2020.

\bibitem[Tian et~al.(2022)Tian, Zhu, Liu, and Zhou]{tian2022image}
Huan Tian, Tianqing Zhu, Wei Liu, and Wanlei Zhou.
\newblock Image fairness in deep learning: problems, models, and challenges.
\newblock \emph{Neural Computing and Applications}, 34\penalty0 (15):\penalty0 12875--12893, 2022.

\bibitem[Torralba and Efros(2011)]{torralba2011unbiased}
Antonio Torralba and Alexei~A Efros.
\newblock Unbiased look at dataset bias.
\newblock In \emph{CVPR 2011}, pages 1521--1528. IEEE, 2011.

\bibitem[Van~Miltenburg(2016)]{van2016stereotyping}
Emiel Van~Miltenburg.
\newblock Stereotyping and bias in the flickr30k dataset.
\newblock \emph{arXiv preprint arXiv:1605.06083}, 2016.

\bibitem[Wang and Russakovsky(2023)]{wang2023overwriting}
Angelina Wang and Olga Russakovsky.
\newblock Overwriting pretrained bias with finetuning data.
\newblock In \emph{Proceedings of the IEEE/CVF International Conference on Computer Vision}, pages 3957--3968, 2023.

\bibitem[Wang et~al.(2022)Wang, Barocas, Laird, and Wallach]{wang2022measuring}
Angelina Wang, Solon Barocas, Kristen Laird, and Hanna Wallach.
\newblock Measuring representational harms in image captioning.
\newblock In \emph{Proceedings of the 2022 ACM Conference on Fairness, Accountability, and Transparency}, pages 324--335, 2022.

\bibitem[Wang et~al.(2019)Wang, Yang, Abdul, and Lim]{wang2019designing}
Danding Wang, Qian Yang, Ashraf Abdul, and Brian~Y Lim.
\newblock Designing theory-driven user-centric explainable ai.
\newblock In \emph{Proceedings of the 2019 CHI conference on human factors in computing systems}, pages 1--15, 2019.

\bibitem[Wang et~al.(2024)Wang, Sun, Wang, Zhang, and Yang]{wang2024navigate}
Yining Wang, Junjie Sun, Chenyue Wang, Mi Zhang, and Min Yang.
\newblock Navigate beyond shortcuts: Debiased learning through the lens of neural collapse.
\newblock In \emph{Proceedings of the IEEE/CVF Conference on Computer Vision and Pattern Recognition}, pages 12322--12331, 2024.

\bibitem[Wang et~al.(2020)Wang, Qinami, Karakozis, Genova, Nair, Hata, and Russakovsky]{wang2020towards}
Zeyu Wang, Klint Qinami, Ioannis~Christos Karakozis, Kyle Genova, Prem Nair, Kenji Hata, and Olga Russakovsky.
\newblock Towards fairness in visual recognition: Effective strategies for bias mitigation.
\newblock In \emph{Proceedings of the IEEE/CVF conference on computer vision and pattern recognition}, pages 8919--8928, 2020.

\bibitem[Wu et~al.(2023)Wu, Yuksekgonul, Zhang, and Zou]{wu2023discover}
Shirley Wu, Mert Yuksekgonul, Linjun Zhang, and James Zou.
\newblock Discover and cure: Concept-aware mitigation of spurious correlation.
\newblock \emph{arXiv preprint arXiv:2305.00650}, 2023.

\bibitem[Wu et~al.(2022)Wu, Xiao, Sun, Zhang, Ma, and He]{wu2022survey}
Xingjiao Wu, Luwei Xiao, Yixuan Sun, Junhang Zhang, Tianlong Ma, and Liang He.
\newblock A survey of human-in-the-loop for machine learning.
\newblock \emph{Future Generation Computer Systems}, 135:\penalty0 364--381, 2022.

\bibitem[Yu et~al.(2022)Yu, Zhong, Qin, Yao, Wang, and He]{yu2022automated}
Boxi Yu, Zhiqing Zhong, Xinran Qin, Jiayi Yao, Yuancheng Wang, and Pinjia He.
\newblock Automated testing of image captioning systems.
\newblock In \emph{Proceedings of the 31st ACM SIGSOFT International Symposium on Software Testing and Analysis}, pages 467--479, 2022.

\bibitem[Zhang et~al.(2022{\natexlab{a}})Zhang, Kuang, Chen, Liu, Wu, and Xiao]{zhang2022fairness}
Fengda Zhang, Kun Kuang, Long Chen, Yuxuan Liu, Chao Wu, and Jun Xiao.
\newblock Fairness-aware contrastive learning with partially annotated sensitive attributes.
\newblock In \emph{The Eleventh International Conference on Learning Representations}, 2022{\natexlab{a}}.

\bibitem[Zhang et~al.(2022{\natexlab{b}})Zhang, Sohoni, Zhang, Finn, and R{\'e}]{zhang2022correct}
Michael Zhang, Nimit~S Sohoni, Hongyang~R Zhang, Chelsea Finn, and Christopher R{\'e}.
\newblock Correct-n-contrast: A contrastive approach for improving robustness to spurious correlations.
\newblock \emph{arXiv preprint arXiv:2203.01517}, 2022{\natexlab{b}}.

\bibitem[Zhang et~al.(2023)Zhang, HaoChen, Huang, Wang, Zou, and Yeung]{zhang2023diagnosing}
Yuhui Zhang, Jeff~Z HaoChen, Shih-Cheng Huang, Kuan-Chieh Wang, James Zou, and Serena Yeung.
\newblock Diagnosing and rectifying vision models using language.
\newblock \emph{arXiv preprint arXiv:2302.04269}, 2023.

\bibitem[Zhang et~al.(2024)Zhang, Feng, Li, and Xu]{zhang2024discover}
Zeliang Zhang, Mingqian Feng, Zhiheng Li, and Chenliang Xu.
\newblock Discover and mitigate multiple biased subgroups in image classifiers.
\newblock In \emph{Proceedings of the IEEE/CVF Conference on Computer Vision and Pattern Recognition}, pages 10906--10915, 2024.

\bibitem[Zhao et~al.(2017)Zhao, Wang, Yatskar, Ordonez, and Chang]{zhao2017men}
Jieyu Zhao, Tianlu Wang, Mark Yatskar, Vicente Ordonez, and Kai-Wei Chang.
\newblock Men also like shopping: Reducing gender bias amplification using corpus-level constraints.
\newblock \emph{arXiv preprint arXiv:1707.09457}, 2017.

\end{thebibliography}
}

\clearpage
\setcounter{page}{1}
\maketitlesupplementary

\section{Implementation}
\subsection{Augmentations}
% this paragraph is a good candidate for moving some or all of to supplementary if we need more space
The image augmentations used in CSBD, which are implemented alongside the weighted sampling for bias mitigation, are also applied to the other methods for fair comparison. On CelebA we use the following augmentations:
\begin{itemize} \item random resized cropping to 224 $\times$ 224;
\item horizontal flips;
\item color jitter;\end{itemize}
and on MS-COCO we use
\begin{itemize}
\item random resized cropping to 448 $\times$ 448;
\item horizontal flips;
\item RandAugment.
\end{itemize}
\subsection{Training and Testing}
The proposed CSBD bias modeling method is CPU-only and does not use any model training. For bias mitigation experiments on downstream tasks, the following training setups are consistent among comparison methods: We train all classification models for 100 epoches and select the best model on validation set to report its results on testing set. The classification model is ResNet50~\cite{he2016deep}. SGD optimizer is used with learning rate of $1e^{-4}$, momentum of 0.9, and weight decay of $0.01$. The batch size is set to be 256 on CelebA dataset and 32 on MS-COCO dataset. For all other method-specific hyper-parameters, we use the default values in the official repositories for LfF,\footnote{https://github.com/alinlab/LfF/} DebiAN,\footnote{https://github.com/zhihengli-UR/DebiAN/} and BPA,\footnote{https://github.com/skynbe/pseudo-attributes} and for PGD we use the default values given in ~\cite{ahn2022mitigating}. These experiments were developed using Pytorch and were performed on a Nvidia Tesla V100 GPU.

\subsection{Algorithm}
\begin{algorithm}[ht]
% \resizebox{0.5\textwidth}{!}{%
% \begin{minipage}{0.5\textwidth}
% \footnotesize
\newcommand{\vect}{\ensuremath{\mathbf{##1}}}

\text{\break}

\textbf{Input:} Image corpus $G = \{g_1, g_2, ..., g_N\}$. \\
\quad \quad \quad Description corpus $D = \{d_1, d_2, ..., d_N\}$ \\

\textbf{Chunking and encoding:}   \\

\quad $D \rightarrow P = \{p_1, p_2, ..., p_N\}$ \\
\quad  \quad \quad \quad \quad \quad \quad \quad \quad \quad \quad $\triangleright$ $P$ is the noun chunk set  \\

\quad $\boldsymbol{A} = e(P)  = \{ \boldsymbol{p}_1, \boldsymbol{p}_2, ..., \boldsymbol{p}_N \} \in \mathbb{R}^{N\times 512} $  \\

\quad \quad \quad \quad $\triangleright$ $e$ is a pre-trained text embedding model \\

\textbf{Agglomerative clustering:}   \\

\quad  $ \{ \boldsymbol{A}, z \} \rightarrow \{C_1, ..., C_F\} $, where $p_i \in C_j$

\quad $\triangleright$ $C$ is a set of noun chunks whose embeddings are clustered together, viewed as a feature \\

% \quad  $\boldsymbol{A} \rightarrow C \text{ sets } \{\boldsymbol{S}_1, ..., \boldsymbol{S}_C\}$ \text{\break} \text{\break} \quad \quad \quad \quad $\triangleright$  Vector cluster \\

% \quad  $A \rightarrow \{S_1, ..., S_C\}$ \quad 
%  $\triangleright$ Natural language phrase cluster  \\

% \quad \textbf{for} $c$ from $1$ to $C$ \textbf{do}: \\

% \quad \quad $\boldsymbol{S_c} \rightarrow F_c \text{ sets } \{\boldsymbol{B}_1, ..., \boldsymbol{B}_{F_c}\}$, s.t. $\text{Var } \boldsymbol{B}_i \le {\sigma}_{max}$ \\

% \quad \quad $S_c \rightarrow \{B_1, ..., B_{F_c}\} $ \\

% \quad  $F = F_1 + F_2 + .. + F_C$ \text{\break} \text{\break} $\triangleright$  Number of feature clusters \\

\textbf{Computing correlation:}   \\

\quad \textbf{for} $C_f$ from $C_1$ to $C_F$ \textbf{do}: \\

\quad \quad $\boldsymbol{t}_f = \{t_1, t_2, ..., t_M\}$, where $t_i = 1$ , if $ \exists \text{\break} p \in C_f $:   $p $ in $d_i$, else $t_i = 0$ \\
\quad \quad \quad \quad \text{\break} \text{\break} $\triangleright$  Binary presence indicator for feature $C_f$\\

\quad \quad 

\quad \textbf{for} $f$ from $1$ to $F$ \textbf{do}: \\

\quad \quad \textbf{for} $f'$ from $1$ to $F$, $f' \neq f$ \textbf{do}: \\

\quad \quad \quad $\boldsymbol{t}_{f}, \boldsymbol{t}_{f'} \rightarrow \phi_{\boldsymbol{t}_{f}\boldsymbol{t}_{f'}} $  \quad $\triangleright$  $\phi$ correlation coefficient \\

\textbf{Outputs:} $[ C_{f}, C_{f'}, \phi_{\boldsymbol{t}_{f}\boldsymbol{t}_{f'}} ]$ , if $ \lvert \phi_{\boldsymbol{t}_{f}\boldsymbol{t}_{f'}} \lvert > \beta $ 

\text{\break}
% \captionsetup{width=\textwidth} % Match caption width to text width
\caption{CSBD. $M$ is the size of the dataset. $N$ is the total number of extracted noun phrase chunks. $z$ is the distance threshold at or above which clusters will not be merged. $F$ is the number of generated clusters or common sense features. $\beta$ is the threshold of feature correlation coefficient.} 
\label{algo:b}
% \end{minipage}
% }
\end{algorithm}

The detailed steps of CSBD method are provided in Algorithm~\ref{algo:b}.

\section{Results - feature correlation list}

We attach the full list of correlated common sense features discovered by our method, on CelebA dataset (Table~\ref{tab:celeba_corr}), and MS-COCO dataset (Table~\ref{tab:coco_corr1} - Table~\ref{tab:coco_corr3}) using dataset captions. It can be seen that many correlations are benign (that is, that the two features' co-occurrence is expected), which can only be judged from human knowledge. Therefore, we maintain a human-in-the-loop component to select spurious correlations from the lists to treat and mitigate downstream model bias.

\begin{table}
\small
\begin{center}
\setlength{\tabcolsep}{4.2pt}
\scalebox{0.78}{
\begin{tabular}{ |c|c|c|c|c| } 
 \hline
  Feature 1 & Feature 2 & $\phi$ correlation coefficient    \\ 
  \hline
 ``\texttt{thick/thin frame}'' & ``\texttt{eyeglasses}'' & 0.1243  \\ 
 \hline
 ``\texttt{thick/thin frame}'' & ``\texttt{no eyeglasses}'' & -0.1328  \\ 
 \hline
 ``\texttt{eyeglasses}'' & ``\texttt{man}'' & 0.0693  \\ 
 \hline
 ``\texttt{eyeglasses}'' & ``\texttt{woman}'' & -0.0743  \\ 
 \hline
 ``\texttt{no eyeglasses}'' & ``\texttt{man}'' & -0.0681 \\ 
 \hline
 ``\texttt{no eyeglasses}'' & ``\texttt{woman}'' & 0.0734  \\ 
 \hline
 ``\texttt{forehead}'' & ``\texttt{eyeglasses}'' & 0.0609  \\ 
 \hline
 ``\texttt{forehead}'' & ``\texttt{a small portion}'' & 0.3503  \\ 
 \hline
 ``\texttt{the face}'' & ``\texttt{no eyeglasses}'' & -0.1849  \\ 
 \hline
 ``\texttt{the face}'' & ``\texttt{eyeglasses}'' & 0.1862  \\ 
 \hline
 ``\texttt{the face}'' & ``\texttt{no smile}'' & 0.0802  \\ 
 \hline
 ``\texttt{the face}'' & ``\texttt{a smile}'' & -0.1012  \\ 
 \hline
 ``\texttt{the face}'' & ``\texttt{happiness}'' & 0.1261  \\ 
 \hline
 ``\texttt{teeth visible}'' & ``\texttt{no smile}'' & -0.3348  \\ 
 \hline
 ``\texttt{teeth visible}'' & ``\texttt{a smile}'' & 0.2571  \\ 
 \hline
 ``\texttt{mouth wide open}'' & ``\texttt{no smile}'' & -0.1826 \\ 
 \hline
 ``\texttt{mouth wide open}'' & ``\texttt{happiness}'' & -0.0934 \\ 
 \hline
 ``\texttt{mouth wide open}'' & ``\texttt{teeth visible}'' & 0.1202 \\ 
 \hline
 ``\texttt{a little bit open}'' & ``\texttt{the face}'' & -0.068 \\ 
 \hline
 ``\texttt{forehead}'' & ``\texttt{man}'' & -0.0637  \\ 
 \hline
 ``\texttt{forehead}''  & ``\texttt{woman}''  & 0.0649\\ 
 \hline
 ``\texttt{no smile}'' & ``\texttt{happiness}'' & -0.2961  \\ 
 \hline
 ``\texttt{no smile}'' & ``\texttt{teenager}'' & 0.0506  \\ 
 \hline
 ``\texttt{no smile}'' & ``\texttt{man}'' & 0.1058  \\ 
 \hline
 ``\texttt{a smile}'' & ``\texttt{man}'' & -0.0552  \\ 
 \hline
 ``\texttt{man}'' & ``\texttt{the middle age}'' & 0.0817  \\ 
 \hline
 ``\texttt{man}'' & ``\texttt{medium length}'' & 0.2204  \\ 
 \hline
 ``\texttt{man}'' & ``\texttt{forties}'', ``\texttt{fifties}'' & 0.0627  \\ 
 \hline
 ``\texttt{beard}'' & ``\texttt{the middle age}'' & 0.056  \\ 
 \hline
 ``\texttt{mustache}'' & ``\texttt{the middle age}'' & 0.0601  \\ 
 \hline
 ``\texttt{beard}'' & ``\texttt{ middle length}'' & 0.18  \\ 
 \hline
 ``\texttt{mustache}'' & ``\texttt{ middle length}'' & 0.1921 \\ 
 \hline
 ``\texttt{beard}'' & ``\texttt{teenager}'' & -0.0504  \\ 
 \hline
 ``\texttt{mustache}'' & ``\texttt{ teenager}'' & -0.0527 \\ 
 \hline
 ``\texttt{beard}'' & ``\texttt{no smile}'' & 0.0754  \\ 
 \hline
 ``\texttt{mustache}'' & ``\texttt{ no smile}'' & 0.0788 \\ 
 \hline
 ``\texttt{beard}'' & ``\texttt{woman}'' & -0.3988  \\ 
 \hline
 ``\texttt{mustache}'' & ``\texttt{woman}'' & -0.3939 \\ 
 \hline
 ``\texttt{beard}'' & ``\texttt{man}'' & 0.4498  \\ 
 \hline
 ``\texttt{mustache}'' & ``\texttt{man}'' & 0.4482 \\ 
 \hline
 
\end{tabular}
}
\end{center}
\caption{\label{tab:celeba_corr} The correlated feature pairs in CelebA dataset, reasoned by CSBD. Each feature is a cluster of descriptive phrases and only 1-2 representative phrases are shown in the table. } 
\end{table}

\begin{table}
\small
\begin{center}
\setlength{\tabcolsep}{4.2pt}
\scalebox{0.78}{
\begin{tabular}{ |c|c|c|c|c| } 
 \hline
  Feature 1 & Feature 2 & $\phi$ correlation coefficient    \\ 
 \hline
 ``\texttt{soccer}'', ``\texttt{baseball}'' & ``\texttt{field}'' & 0.1246 \\ 
 \hline
 ``\texttt{baseball bat}'' & ``\texttt{field}'' & 0.0902 \\ 
 \hline
 ``\texttt{soccer}'', ``\texttt{baseball}'' & ``\texttt{the ball}'' & 0.1638  \\ 
 \hline
 ``\texttt{soccer}'', ``\texttt{baseball}'' & ``\texttt{a game}'' & 0.149  \\ 
 \hline
 ``\texttt{baseball bat}''  & ``\texttt{a ball}''  & 0.1089 \\ 
 \hline
 ``\texttt{soccer}'', ``\texttt{baseball}'' & ``\texttt{a man}'', ``\texttt{a boy}'' & 0.0673  \\ 
 \hline
 ``\texttt{sky}'', ``\texttt{ air}'' & ``\texttt{a man}'', ``\texttt{a boy}'' & 0.0558  \\ 
 \hline
 ``\texttt{sky}'', ``\texttt{ air}'' &  ``\texttt{skateboard}'' & 0.1252  \\ 
 \hline
 ``\texttt{sky}'', ``\texttt{ air}'' &  ``\texttt{snowboard}'' & 0.0903  \\  
 \hline
 ``\texttt{sky}'', ``\texttt{ air}'' &  ``\texttt{airplane}'' & 0.0901 \\ 
 \hline
 ``\texttt{sky}'', ``\texttt{ air}'' &  ``\texttt{kite}'' & 0.1024 \\ 
 \hline
 ``\texttt{men}'' &  ``\texttt{a group}'' & 0.085 \\ 
 \hline
 ``\texttt{a man}'' &  ``\texttt{a group}'' & -0.0666 \\ 
 \hline
 ``\texttt{a man}''&  ``\texttt{frisbee}'' & 0.0701 \\ 
 \hline
 ``\texttt{a man}'' &  ``\texttt{skateboard}'' & 0.1263\\ 
 \hline
 ``\texttt{a man}''&  ``\texttt{snowboard}'' & 0.0885\\ 
 \hline
 ``\texttt{men}'' &  ``\texttt{two members}'' & 0.0768\\ 
 \hline
 ``\texttt{men}'' &  ``\texttt{two members}'' & 0.0768\\ 
 \hline
 ``\texttt{a man}'' &  ``\texttt{cellphone}'' & 0.0523 \\
 \hline
 ``\texttt{a man}'' &  ``\texttt{hand}'', ``\texttt{fingers}'' &  0.051\\ 
 \hline
 ``\texttt{a man}'' &  ``\texttt{tennis racket}'' & 0.0674 \\ 
 \hline
 ``\texttt{a man}'' &  ``\texttt{tennis player}'' & 0.0758 \\ 
 \hline
 ``\texttt{a man}'' &  ``\texttt{umbrella}'' & 0.062 \\ 
 \hline
 ``\texttt{men}'' &  ``\texttt{sneakers}'', ``\texttt{pants}'' & 0.0621 \\ 
 \hline
 ``\texttt{a man}'' &  ``\texttt{wave}'' & 0.0969 \\ 
 \hline
 ``\texttt{a man}'' &  ``\texttt{surfboard}'' & 0.1102 \\ 
 \hline
 ``\texttt{zebra}'' &  ``\texttt{ grass field}'' & 0.1454 \\ 
 \hline
 ``\texttt{giraffe}'' &  ``\texttt{ grass field}'' & 0.1022 \\ 
 \hline
 ``\texttt{zebra}'' &  ``\texttt{sheep}'' & 0.077 \\ 
 \hline
 ``\texttt{zebra}'' &  ``\texttt{trees}'' & 0.0834 \\ 
 \hline
 ``\texttt{giraffe}'' &  ``\texttt{trees}'' & 0.1644 \\ 
 \hline
 ``\texttt{giraffe}'' &  ``\texttt{fence}'' & 0.0745 \\ 
 \hline
 ``\texttt{a person}'' &  ``\texttt{snowy slopes}'' & 0.084 \\ 
 \hline
 ``\texttt{a person}'' &  ``\texttt{ skateboard}'' & 0.0635 \\ 
 \hline
 ``\texttt{a person}'' &  ``\texttt{snowboard}'' & 0.0748 \\ 
 \hline
 ``\texttt{bike}'' &  ``\texttt{city street}'' & 0.0675 \\ 
 \hline
 ``\texttt{motorcycle}'' &  ``\texttt{city street}'' &  0.0504 \\ 
 \hline
 ``\texttt{motorcycle}'' &  ``\texttt{road}'' &  0.0677 \\ 
 \hline
 ``\texttt{bus}'' &  ``\texttt{city street}'' &  0.1364\\ 
 \hline
 ``\texttt{bus}'' &  ``\texttt{road}'' & 0.1072 \\ 
 \hline
 ``\texttt{bathroom}'' &  ``\texttt{ vase}'' & 0.1229 \\ 
 \hline
 ``\texttt{sink}'' &  ``\texttt{window}'' & 0.085 \\ 
 \hline
 ``\texttt{bathroom}'' &  ``\texttt{window}'' & 0.1092 \\ 
 \hline
 ``\texttt{sink}'' &  ``\texttt{toilet}'' & 0.1667 \\ 
 \hline
 ``\texttt{bathroom}'' &  ``\texttt{toilet}'' & 0.423 \\ 
 \hline
 ``\texttt{snowy slopes}'' &  ``\texttt{snowboard}'' & 0.1701 \\ 
 \hline
 ``\texttt{table}'' &  ``\texttt{food}'' & 0.1723 \\ 
 \hline
 ``\texttt{table}'' &  ``\texttt{pizza}'' & 0.1123 \\ 
 \hline
 ``\texttt{table}'' &  ``\texttt{cake}'' & 0.081 \\ 
 \hline
 ``\texttt{table}'' &  ``\texttt{beverages}'' &  0.1173 \\ 
 \hline
 ``\texttt{table}'' &  ``\texttt{flower}'' & 0.0642 \\ 
 \hline
 ``\texttt{table}'' &  ``\texttt{plate}'' & 0.123 \\ 
 \hline
 ``\texttt{table}'' &  ``\texttt{bowl}'', `\texttt{cup}'' & 0697 \\ 
 \hline
 ``\texttt{table}'' &  ``\texttt{fresh produce}'' & 0.0634 \\ 
 \hline
 ``\texttt{table}'' &  ``\texttt{vase}'' & 0.079 \\ 
 \hline
 ``\texttt{table}'' &  ``\texttt{glasses}'' & 0.0578 \\ 
 \hline
 ``\texttt{food}'' &  ``\texttt{kitchen}'' & 0.0703 \\ 
 \hline
 ``\texttt{food}'' &  ``\texttt{plate}'' &  0.2588 \\ 
 \hline
 ``\texttt{food}'' &  ``\texttt{a dish}'' & 0.0693 \\ 
 \hline
 ``\texttt{food}'' &  ``\texttt{bowl}'', `\texttt{cup}'' & 0.0516 \\ 
 \hline
 ``\texttt{food}'' &  ``\texttt{egg}'', ``\texttt{sausage}'' &  0.0545 \\ 
 \hline
 ``\texttt{living room}'' &  ``\texttt{game controller}'' &  0.0574 \\ 
 \hline
 ``\texttt{living room}'' &  ``\texttt{pillow}'' & 0.1643 \\ 
 \hline
 ``\texttt{couch}'' &  ``\texttt{dog}'' &  0.058 \\ 
 \hline
 ``\texttt{couch}'' &  ``\texttt{cat}'' &  0.0615 \\ 
 \hline
\end{tabular}
}
\end{center}
\caption{\label{tab:coco_corr1} The correlated feature pairs in MS-COCO dataset, reasoned by CSBD. Each feature is a cluster of descriptive phrases and only 1-2 representative phrases are shown in the table.} 
\end{table}

\begin{table}
\small
\begin{center}
\setlength{\tabcolsep}{4.2pt}
\scalebox{0.78}{
\begin{tabular}{ |c|c|c|c|c| } 
 \hline
  Feature 1 & Feature 2 & $\phi$ correlation coefficient    \\ 
  \hline
  ``\texttt{the top}'', ``\texttt{the front}'' &  ``\texttt{clock}'' & 0.0594  \\ 
 \hline
 ``\texttt{the top}'', ``\texttt{the front}'' &  ``\texttt{car}'' & 0.0543  \\ 
 \hline
 ``\texttt{microwave}'', ``\texttt{oven}'' &  ``\texttt{dishes}'' & 0.0604  \\ 
 \hline
 ``\texttt{kitchen}'' &  ``\texttt{scissors}'' & 0.0894  \\ 
 \hline
 ``\texttt{a group}'' &  ``\texttt{a game}'' & 0.0588  \\ 
 \hline
 ``\texttt{a group}'' &  ``\texttt{people}'' &  0.3927 \\ 
 \hline
 ``\texttt{frisbee}'' &  ``\texttt{field}'' & 0.0616  \\ 
 \hline
 ``\texttt{frisbee}'' &  ``\texttt{a game}'' & 0.0994  \\ 
 \hline
 ``\texttt{frisbee}'' &  ``\texttt{dog}'' &  0.091 \\ 
 \hline
 ``\texttt{plate}'' &  ``\texttt{pizza}'' & 0.1148  \\ 
 \hline
 ``\texttt{plate}'' &  ``\texttt{cake}'' & 0.1013  \\ 
 \hline
 ``\texttt{plate}'' &  ``\texttt{sandwich}'' & 0.2136  \\ 
 \hline
 ``\texttt{plate}'' &  ``\texttt{donuts}'' & 0.0614  \\ 
 \hline
 ``\texttt{plate}'' &  ``\texttt{meat}'' & 0.1573  \\ 
 \hline
 ``\texttt{plate}'' &  ``\texttt{vegetables}'' & 0.1855  \\ 
 \hline
 ``\texttt{plate}'' &  ``\texttt{snacks}'' & 0.1307  \\ 
 \hline
 ``\texttt{plate}'' &  ``\texttt{hot dogs}'' & 0.0594  \\ 
 \hline
 ``\texttt{plate}'' &  ``\texttt{fruits}'' & 0.074  \\ 
 \hline
 ``\texttt{plate}'' &  ``\texttt{beverages}'' & 0.0627  \\ 
 \hline
 ``\texttt{bowl}'', ``\texttt{cup}'' &  ``\texttt{sandwich}'' & 0.069  \\
 \hline
 ``\texttt{bowl}'', ``\texttt{cup}'' &  ``\texttt{beverages}'' & 0.1497  \\
 \hline
 ``\texttt{bowl}'', ``\texttt{cup}'' &  ``\texttt{fruit}'' & 0.1396  \\
 \hline
 ``\texttt{bowl}'', ``\texttt{cup}'' &  ``\texttt{vegetables}'' & 0.1079  \\
 \hline
 ``\texttt{bowl}'', ``\texttt{cup}'' &  ``\texttt{bananas}'' & 0.0625  \\
 \hline
 ``\texttt{glasses}'' &  ``\texttt{beverages}'' & 0.136  \\ 
 \hline
 ``\texttt{a pot}'' &  ``\texttt{beverages}'' & 0.0942  \\ 
 \hline
 ``\texttt{a pot}'' &  ``\texttt{flower}'' & 0.2277  \\ 
 \hline
 ``\texttt{sandwich}'' &  ``\texttt{beverages}'' & 0.0537  \\
 \hline
 ``\texttt{sandwich}'' &  ``\texttt{meat}'' & 0.0725  \\ 
 \hline
 ``\texttt{sandwich}'' &  ``\texttt{vegetables}'' & 0.0812  \\ 
 \hline
 ``\texttt{hot dogs}'' &  ``\texttt{vegetables}'' & 0.1717  \\ 
 \hline
 ``\texttt{donuts}'' &  ``\texttt{beverages}'' & 0.0515  \\ 
 \hline
 ``\texttt{pizza}'' &  ``\texttt{vegetables}'' & 0.1046  \\ 
 \hline
 ``\texttt{pizza}'' &  ``\texttt{box}'' & 0.0784  \\ 
 \hline
 ``\texttt{cake}'' &  ``\texttt{box}'' & 0.1151  \\ 
 \hline
 ``\texttt{donuts}'' &  ``\texttt{box}'' & 0.0664  \\ 
 \hline
 ``\texttt{tray}'' &  ``\texttt{pizza}'' & 0.1309  \\ 
 \hline
 ``\texttt{tray}'' &  ``\texttt{cake}'' & 0.0529  \\ 
 \hline
 ``\texttt{tray}'' &  ``\texttt{sandwich}'' & 0.0586  \\ 
 \hline
 ``\texttt{tray}'' &  ``\texttt{vegetables}'' & 0.105  \\ 
 \hline
 ``\texttt{dish}'' &  ``\texttt{vegetables}'' & 0.0836  \\ 
 \hline
 ``\texttt{skateboard}'' &  ``\texttt{a line drive}'' &  0.0849 \\ 
 \hline
 ``\texttt{skateboard}'' &  ``\texttt{boys}'' & 0.1404  \\ 
 \hline
 ``\texttt{grass field}'' &  ``\texttt{cow}'' &  0.1226 \\ 
 \hline
 ``\texttt{grass field}'' &  ``\texttt{sheep}'' & 0.184  \\ 
 \hline
 ``\texttt{grass field}'' &  ``\texttt{elephant}'' & 0.0516  \\ 
 \hline
 ``\texttt{field}'' &  ``\texttt{trees}'' & 0.063  \\ 
 \hline
 ``\texttt{sheep}'' &  ``\texttt{birds}'' & 0.0794  \\ 
 \hline
 ``\texttt{bed}'' &  ``\texttt{chairs}'' & 0.051  \\ 
 \hline
 ``\texttt{bedroom}'' &  ``\texttt{chairs}'' &  0.057 \\ 
 \hline
 ``\texttt{bedroom}'' &  ``\texttt{furniture}'' & 0.0852  \\ 
 \hline
 ``\texttt{bed}'' &  ``\texttt{cat}'' &  0.0609 \\ 
 \hline
 ``\texttt{bags}'' &  ``\texttt{a pack}'' & 0.0506  \\ 
 \hline
 ``\texttt{city}'' &  ``\texttt{traffic lights}'' &  0.0682 \\ 
 \hline
 ``\texttt{cell phone}'' &  ``\texttt{women}'' & 0.0721  \\ 
 \hline
 ``\texttt{desk}'' &  ``\texttt{cat}'' &  0.0753 \\ 
 \hline
 ``\texttt{keyboard}'' &  ``\texttt{cat}'' &  0.1 \\ 
 \hline
 ``\texttt{beach}'' &  ``\texttt{kite}'' & 0.1347  \\ 
 \hline
 ``\texttt{beach}'' &  ``\texttt{people}'' & 0.0789  \\ 
 \hline
 ``\texttt{beach}'' &  ``\texttt{surfer}'' & 0.1594  \\ 
 \hline
  ``\texttt{beach}'' &  ``\texttt{umbrella}'' & 0.0768  \\ 
 \hline
 ``\texttt{water}'', ``\texttt{ocean}'' &  ``\texttt{boat}'' &  0.1235 \\ 
 \hline
 ``\texttt{water}'', ``\texttt{ocean}'' &  ``\texttt{birds}'' &  0.0558 \\ 
 \hline
 ``\texttt{water}'', ``\texttt{ocean}'' &  ``\texttt{shore}'' &  0.1412 \\ 
 \hline
 ``\texttt{water}'', ``\texttt{ocean}'' &  ``\texttt{surfer}'' &  0.2677 \\ 
 \hline
\end{tabular}
}
\end{center}
\caption{\label{tab:coco_corr2} The correlated feature pairs in MS-COCO dataset, reasoned by CSBD. Each ``feature'' is a cluster of descriptive noun chunks and only 1-2 representative noun chunks are shown in the table.} 
\end{table}

\begin{table}
\small
\begin{center}
\setlength{\tabcolsep}{4.3pt}
\scalebox{0.8}{
\begin{tabular}{ |c|c|c|c|c| } 
 \hline
  Feature 1 & Feature 2 & $\phi$ correlation coefficient    \\ 
  \hline
  ``\texttt{a woman}'' &  ``\texttt{tennis player}'' & 0.0902  \\ 
  \hline
  ``\texttt{a woman}'' & ``\texttt{umbrella}'' & 0.0688  \\ 
  \hline
  ``\texttt{city street}'' & ``\texttt{sign}'' &  0.0743 \\ 
  \hline
  ``\texttt{city street}'' &``\texttt{traffic light}'' & 0.1255  \\ 
  \hline
  ``\texttt{fruit}'' & ``\texttt{a bunch}'' & 0.0644  \\ 
  \hline
  ``\texttt{bananas}'' & ``\texttt{a bunch}'' & 0.0938  \\ 
  \hline
  ``\texttt{clock}'' & ``\texttt{building}'' &  0.1026 \\ 
  \hline
  ``\texttt{airplane}'' &``\texttt{airport}'' & 0.1426  \\ 
  \hline
  ``\texttt{trees}'' & ``\texttt{elephant}'' & 0.0637  \\ 
  \hline
  ``\texttt{a ball}'' & ``\texttt{tennis player}'' &  0.1709 \\ 
  \hline
  ``\texttt{a ball}'' & ``\texttt{tennis racket}'' & 0.0833  \\ 
  \hline
  ``\texttt{a ball}'' & ``\texttt{tennis court}'' & 0.0827  \\ 
  \hline
  ``\texttt{tennis court}'' & ``\texttt{a game}'' &  0.0595 \\ 
  \hline
  ``\texttt{a game}'' & ``\texttt{controllers}'' & 0.1199  \\ 
  \hline
  ``\texttt{a game}'' & ``\texttt{a man}'' &  0.0642 \\ 
  \hline
  ``\texttt{window}'' & ``\texttt{cat }'' &  0.0574 \\ 
  \hline
  ``\texttt{building}'' & ``\texttt{sign}'' & 0.056  \\ 
  \hline
  ``\texttt{a flattened area}'' & ``\texttt{train}'' & 0.093  \\ 
  \hline
  ``\texttt{boat}'' & ``\texttt{water flow}'' & 0.0802  \\ 
  \hline
  ``\texttt{fence}'' & ``\texttt{sign}'' & 0.1064  \\ 
  \hline
  ``\texttt{umbrella}'' & ``\texttt{people}'' & 0.1001  \\ 
  \hline
  ``\texttt{kite}'' & ``\texttt{people}'' & 0.1023  \\ 
  \hline
  ``\texttt{birds}'' & ``\texttt{water flow}'' & 0.0522  \\ 
  \hline
  ``\texttt{people}'' & ``\texttt{a crowd}'' & 0.097  \\ 
  \hline
\end{tabular}
}
\end{center}
\caption{\label{tab:coco_corr3} The correlated feature pairs in MS-COCO dataset, reasoned by CSBD. Each ``feature'' is a cluster of descriptive noun chunks and only 1-2 representative noun chunks are shown in the table.} 
\end{table}

% \section{Rationale}
% \label{sec:rationale}
% % 
% Having the supplementary compiled together with the main paper means that:
% % 
% \begin{itemize}
% \item The supplementary can back-reference sections of the main paper, for example, we can refer to \cref{sec:intro};
% \item The main paper can forward reference sub-sections within the supplementary explicitly (e.g. referring to a particular experiment); 
% \item When submitted to arXiv, the supplementary will already included at the end of the paper.
% \end{itemize}
% % 
% To split the supplementary pages from the main paper, you can use \href{https://support.apple.com/en-ca/guide/preview/prvw11793/mac#:~:text=Delete%20a%20page%20from%20a,or%20choose%20Edit%20%3E%20Delete).}{Preview (on macOS)}, \href{https://www.adobe.com/acrobat/how-to/delete-pages-from-pdf.html#:~:text=Choose%20%E2%80%9CTools%E2%80%9D%20%3E%20%E2%80%9COrganize,or%20pages%20from%20the%20file.}{Adobe Acrobat} (on all OSs), as well as \href{https://superuser.com/questions/517986/is-it-possible-to-delete-some-pages-of-a-pdf-document}{command line tools}.

\end{document}